\documentclass[11pt]{article}
\usepackage[preprint]{acl}
\usepackage{times}
\usepackage{latexsym}
\usepackage{inconsolata} 
\usepackage[T1]{fontenc}
\usepackage[utf8]{inputenc}
\usepackage{microtype}

\usepackage{algorithm}
\usepackage{algpseudocode}
\usepackage{amsmath}
\usepackage{amssymb}
\usepackage{amsfonts}
\usepackage{booktabs}
\usepackage{nicefrac}
\usepackage{xcolor}
\usepackage{graphicx}
\usepackage{enumitem}
\usepackage{subcaption}

\usepackage{multirow}
\usepackage{float}

\newcommand{\mask}{\text{[MASK]}}

\usepackage{amsmath,amsfonts,bm}

\def\eqref#1{equation~\ref{#1}}

\def\1{\bm{1}}

\DeclareMathAlphabet{\mathsfit}{\encodingdefault}{\sfdefault}{m}{sl}
\SetMathAlphabet{\mathsfit}{bold}{\encodingdefault}{\sfdefault}{bx}{n}

\title{CreditDecoding: Accelerating Parallel Decoding in Diffusion Large Language Models with Trace Credit}

\author{
  Kangyu Wang\textsuperscript{1,2,*},
  Zhiyun Jiang\textsuperscript{2,3,*},
  Haibo Feng\textsuperscript{2},
  Weijia Zhao\textsuperscript{2} \\
  \textbf{Lin Liu\textsuperscript{2}},
  \textbf{Jianguo Li\textsuperscript{2,$\dagger$}},
  \textbf{Zhenzhong Lan\textsuperscript{2,3,$\dagger$}},
  \textbf{Weiyao Lin\textsuperscript{1,$\dagger$}} \\
  \textsuperscript{1}Shanghai Jiao Tong University, 
  \textsuperscript{2}Ant Group,
  \textsuperscript{3}Westlake University
 }
\begin{document}
\maketitle
\renewcommand{\thefootnote}{\fnsymbol{footnote}}
\footnotetext[1]{Equal Contribution. Work done during internship at Ant Group. Correspondence to \texttt{kangyuwang@sjtu.edu.cn} and \texttt{zhiyunjiang@westlake.edu.cn}.}
\footnotetext[2]{Corresponding authors.}

\begin{abstract}
Diffusion large language models (dLLMs) generate text through iterative denoising. In commonly adopted parallel decoding schemes, each step confirms only high-confidence positions while remasking the others.
By analyzing dLLM denoising traces, we uncover a key inefficiency: models often predict the correct target token several steps before its confidence becomes high enough to be decoded.
This gap between early prediction and late decoding forces repeated remasking of already-correct tokens, causing redundant iterations and limiting acceleration.
To exploit this temporal redundancy, we introduce \emph{Trace Credit} to quantify a token's decoding potential by accumulating historical evidence.
Building on this, we propose \textbf{CreditDecoding}, a \textit{training-free} parallel decoding method that fuses Trace Credit with current logits to boost the confidence of correct but underconfident tokens, thereby accelerating denoising and improving robustness.
On eight benchmarks, CreditDecoding achieves up to 5.48$\times$ speedup with +0.48 accuracy on LLaDA-8B-Instruct, and consistently improves performance across diverse dLLM architectures and parameter scales.
It further scales to long contexts and remains orthogonal to mainstream inference optimizations, making it a practical and applicable solution.
\end{abstract}

\begin{figure}[h] 
    \centering
    \includegraphics[width=\linewidth]{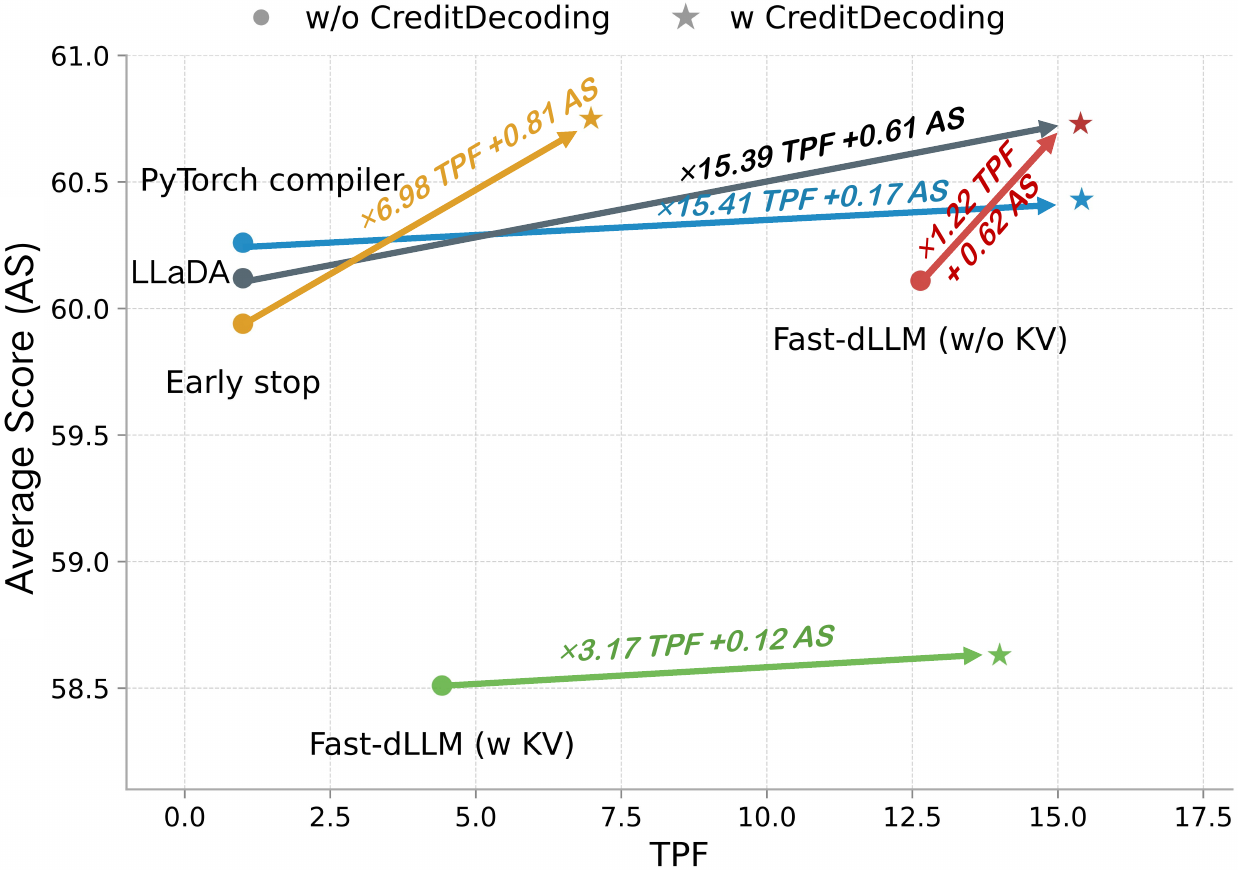}
    \caption{Acceleration methods (dots) vs. their performance with CreditDecoding (stars) on LLaDA.}
    \label{fig:Orthogonality}
\end{figure}

\section{Introduction}
\begin{figure*}[t]
    \centering
    \includegraphics[width=\linewidth]{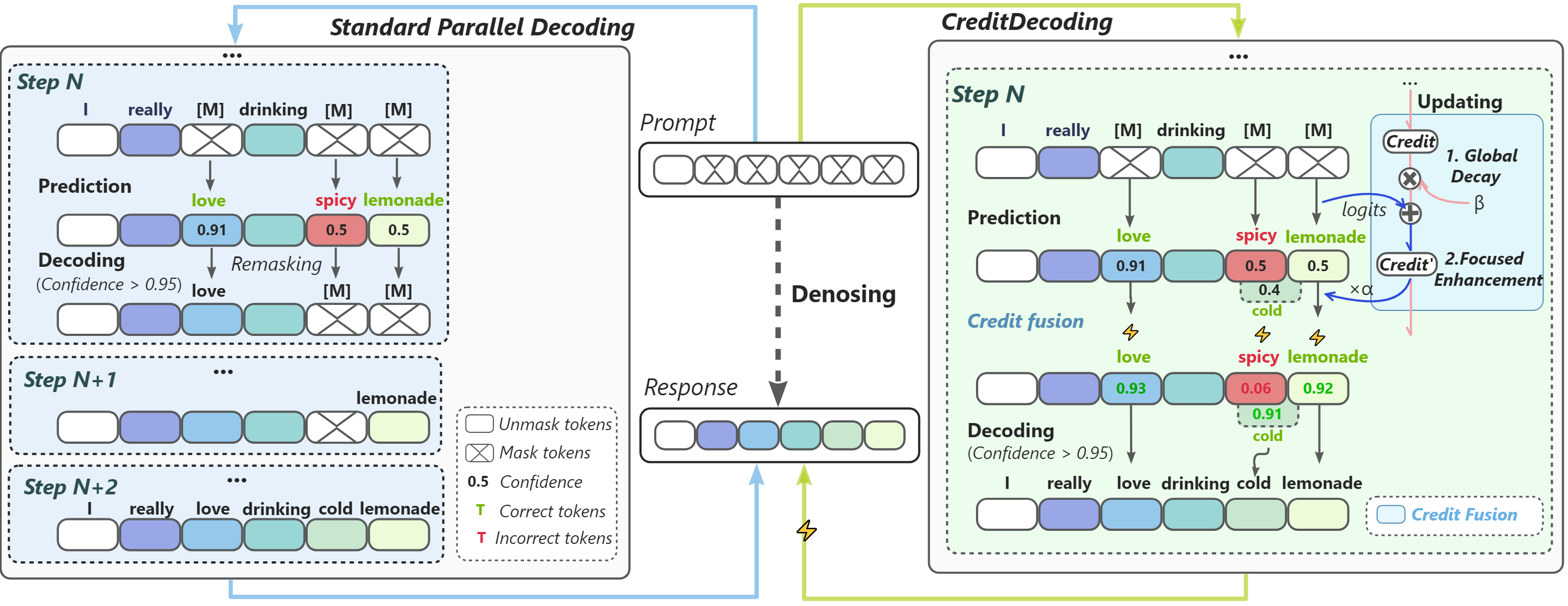}
    \caption{Comparison between \textbf{standard dLLM parallel decoding} (left) and the proposed \textbf{CreditDecoding} (right). The left diagram illustrates how existing methods predict solely on instantaneous predictions at each step, causing the repetitive remasking of correct tokens. In contrast, CreditDecoding maintains a token-level credit value across steps, using \emph{trace credit} as a prior to enhance and calibrate current predictions.}
    \label{fig:diagram}
\end{figure*}
Diffusion-based large language models (dLLMs) have recently emerged as a promising alternative to autoregressive models (ARMs) for text generation \citep{ye2025dream,nie2025large,zhu2025llada,llada-moe-7b-a1b-base,gong2025diffucoderunderstandingimprovingmasked,gong2025scalingdiffusionlanguagemodels,song2025seeddiffusionlargescalediffusion,yang2025mmadamultimodallargediffusion,kim2025anyorderflexiblelengthmasked}. Unlike ARMs that predict tokens strictly left-to-right, dLLMs generate text through iterative denoising with bidirectional attention, enabling richer contextual dependencies. This paradigm has demonstrated advantages in reasoning and generation quality \citep{nie2025large,ye2025dream}. However, inference efficiency remains a major bottleneck: dLLMs typically require redundant denoising steps to predict all masked tokens, and the use of bidirectional attention precludes a lossless KV cache \citep{yu2025dimplediscretediffusionmultimodal}.

To accelerate inference, recent work has focused on improving the effectiveness and efficiency of \emph{parallel decoding} in dLLMs \citep{yu2025dimplediscretediffusionmultimodal,wei2025acceleratingdiffusionlargelanguage}. At each denoising step, the model first predicts all masked tokens and then selects a subset of high-confidence positions to decode, while remasking the remaining uncertain tokens for future refinement \citep{yu2025dimplediscretediffusionmultimodal}. This strategy allows multiple tokens to be updated in parallel and has proven both simple and effective. Nevertheless, it suffers from two key limitations: 

\noindent\textbf{(i) Computational redundancy.}
In many cases, tokens are predicted early but decoded late due to low confidence, causing repeated predictions. Figure~\ref{fig5:rank_correct} shows the resulting temporal gap between initial prediction and final decoding.

\noindent\textbf{(ii) History-agnostic Decoding.} 
As the context is iteratively updated and may include mispredicted tokens, the confidence of otherwise stable tokens can fluctuate or even regress \citep{wang2025timefeatureexploitingtemporal}. 
However, token decoding is typically independent of predictions from previous steps, 
without leveraging the historical consistency of tokens. This undermines convergence and propagates errors across steps, reducing decoding robustness.

To address these issues, we propose \textbf{CreditDecoding}, a \emph{training-free} parallel decoding strategy for dLLMs. It assigns each token a \emph{trace credit} score that accumulates historical logits across steps. This score acts as a prior condition fused with the current logits to boost
confidence. CreditDecoding reduces redundant iterations while stabilizing predictions against temporary inconsistencies.

We evaluate CreditDecoding on four dLLMs across eight benchmarks covering knowledge, reasoning, and coding. 
Experiments show that CreditDecoding achieves consistent speedups with modest performance uplift across a diverse set of dLLM architectures and benchmarks, while remaining orthogonal to mainstream optimizations such as KV cache \citep{feng2025theoreticalbenefitlimitationdiffusion} and kernel fusion \citep{pytorch20}.
In summary, our work makes the following contributions:

\begin{itemize}[noitemsep,topsep=0pt,partopsep=0pt,parsep=0pt]
    \item We analyze threshold-based parallel decoding and identify \emph{computational redundancy} and \emph{history-agnostic decisions} as key bottlenecks. We further find \emph{temporal consistency} in confidence trajectories across denoising steps, which provides an effective prior for acceleration.
    
    \item We propose \textbf{CreditDecoding}, a \emph{training-free} method that accumulates trace credit as a token-level prior to accelerate inference. We also introduce a \emph{tuning-free} variant to improve usability. On LLaDA-8B-Instruct, the primary approach achieves up to $5.48\times$ speedup while \emph{improving} task performance.
    
    \item We demonstrate the \emph{scalability}, \emph{generality}, and \emph{orthogonality} of CreditDecoding. 
    It scales effectively across a wide range of model sizes and context lengths, while maintaining seamless compatibility with mainstream optimizations (Figure~\ref{fig:Orthogonality}).

\end{itemize}

\section{Related Work}

\textbf{Diffusion Language Models }
Diffusion large language models (dLLMs) replace left-to-right prediction with iterative denoising, enabling order-agnostic and parallel token updates with bidirectional context \citep{nie2025large,ye2025dream}. Representative systems include Dream and the LLaDA family, with extensions to code, large-scale training, and multimodal/vision-conditioned settings \citep{ye2025dream,nie2025large,zhu2025llada,llada-moe-7b-a1b-base,gong2025diffucoderunderstandingimprovingmasked,gong2025scalingdiffusionlanguagemodels,song2025seeddiffusionlargescalediffusion,you2025lladavlargelanguagediffusion,yang2025mmadamultimodallargediffusion}. Variants further explore flexible-length and any-order masking \citep{kim2025anyorderflexiblelengthmasked}. Theoretically, dLLMs can approach autoregressive quality but typically require multiple denoising steps, with complexity that may grow with stricter sequence-level correctness and longer context \citep{feng2025theoreticalbenefitlimitationdiffusion,liu2025longlladaunlockinglongcontext}. Practically, unlike ARMs, dLLMs lack lossless KV cache and often incur high latency due to many denoising iterations \citep{cobbe2021trainingverifierssolvemath,hendrycks2021measuringmathematicalproblemsolving,chen2021evaluatinglargelanguagemodels,jain2024livecodebenchholisticcontaminationfree}.

\textbf{Inference Acceleration and Decoding Strategies }
To reduce latency, parallel decoding samples multiple tokens per step, and reusing or caching of bidirectional-attention outputs or other stable computations can further cut runtime without retraining \citep{yu2025dimplediscretediffusionmultimodal,wei2025acceleratingdiffusionlargelanguage,liu2025dllmcacheacceleratingdiffusionlarge}. Beyond speed, planning and ordering methods (e.g., path-planning remasking, adaptive ordering, calibration, or attention pruning) improve robustness and efficiency \citep{peng2025pathplanningmaskeddiffusion,kim2025trainworstplanbest,huang2025pcsamplerpositionawarecalibrationdecoding,chen2025dpadefficientdiffusionlanguage}, and some incorporate temporal/historical signals \citep{wang2025timefeatureexploitingtemporal}. However, many score-based strategies remain largely history-agnostic, relying on current-step confidence, which can induce step-level instability and redundant remasking until confidence converges.

\section{Preliminary}
\subsection{Inference Process of dLLMs}
A Diffusion Large Language Model (dLLM) generates discrete text sequences by iteratively denoising a fully masked input. 
dLLMs formulate generation as a stochastic reverse denoising process that starts from an all-\texttt{[Mask]} sequence and gradually recovers the clean sequence.

Let $x \in \mathcal{V}^L$ be a sequence of length $L$ over a vocabulary $\mathcal V$. At each discrete step $t \in \{0,\dots,T\}$, let $x_t$ denote the corrupted sequence and $M_t \subseteq \{1,\dots,L\}$ be the set of masked positions. We use $\eta_t \in [0,1]$ to denote the masked ratio, following a forward schedule where noise increases with $t$.

The core component is a denoising model $f_\theta$ that maps a corrupted sequence $x_t$ to logits $l_t=f_\theta(x_t)$. These logits parameterize the probability distribution of the original token at each position $i$:
\begin{equation}
\label{eq:prob}
p_\theta^i(\cdot \mid x_t) = \mathrm{Softmax}(l_t^i).
\end{equation}

This order-agnostic formulation enables the model to predict masked tokens based on arbitrary visible context. During inference, the denoising process starts from $x_T$ and iteratively refines the state until the clean sequence $x_0$ is fully recovered. At each denoising step $t$, the transition from $x_t$ to $x_{t-1}$ is defined as:
\begin{equation}
x_{t-1} \sim \prod_{i=1}^L g_\theta(x_{t-1}^i \mid x_t),
\end{equation}
where the transition kernel $g_\theta$ is given by:
\begin{equation}
g_\theta(x_{t-1}^i \mid x_{t})=
\begin{cases}
x_t^i, & i\notin M_t,\\
\mathrm{Cat}\!\big(\pi_t^i\big), & i\in M_t.
\end{cases}
\end{equation}
Here, $\pi_t^i(\cdot)=\tfrac{\eta_{t-1}}{\eta_t}\mathbf{e}_M+\tfrac{\eta_t-\eta_{t-1}}{\eta_t}p_\theta^i(\cdot\mid x_t)$, and $\mathbf{e}_M$ denotes the one-hot vector of the \texttt{[Mask]}.

Intuitively, at each masked position, the token either remains masked or is predicted by the model based on the schedule. In practical parallel decoding, high-confidence tokens from $p_\theta(\cdot|x_t)$ are decoded and updated into $x_{t-1}$, while uncertain positions are remasked for future refinement. This procedure iterates until $M_0 = \varnothing$.

Moreover, many implementations adopt a \textbf{block-wise} strategy: the sequence is partitioned into blocks, and tokens within the current block can be decoded while external tokens serve as fixed context. This limits the impact of uncertain tokens, reducing error propagation and improving stability.

\subsection{Parallel Decoding}
dLLMs naturally support recovering masked tokens in parallel. Assuming conditional independence at each step $t$, the joint distribution is approximated by the product of marginals. 
For each masked position $i\in M_t$, we \textit{greedily sample} the top-$1$ token $\tilde{x}t^{i} = \arg\max{v} p_\theta^i(v\mid x_t)$ and its confidence score $s_t^{i}=p_\theta^i(\tilde{x}_t^{i} \mid x_t)$ from all candidate tokens $v \in \mathcal{V}$.  
  
Previous work validates that high-confidence marginals effectively approximate the joint distribution \citep{wu2025fastdllmtrainingfreeaccelerationdiffusion}. Based on this, the mainstream strategy decodes a subset of tokens $I_t = \{ i \in M_t \mid s_t^i \geq \tau \}$ whose confidence exceeds a threshold $\tau$. Tokens in $I_t$ are decoded in parallel, while the rest remain masked (Figure~\ref{fig:diagram}).

Beyond simple probability, other scoring functions (e.g., negative entropy, probability margins) and sampling schemes (e.g., Top-$K$, adaptive schedules) have also been explored.
\section{Methodology}
\subsection{Observations}

In this section, we analyze limitations of threshold-based parallel decoding in dLLM inference.
\label{definition-of-target-token}
For each position $i$, we define the \textbf{\textit{target token}} $x_0^{i,v}$ as the token with ID $v$ ultimately decoded at position $i$ in the final output sequence.

\begin{figure}[t]
    \centering
    \includegraphics[width=\linewidth]{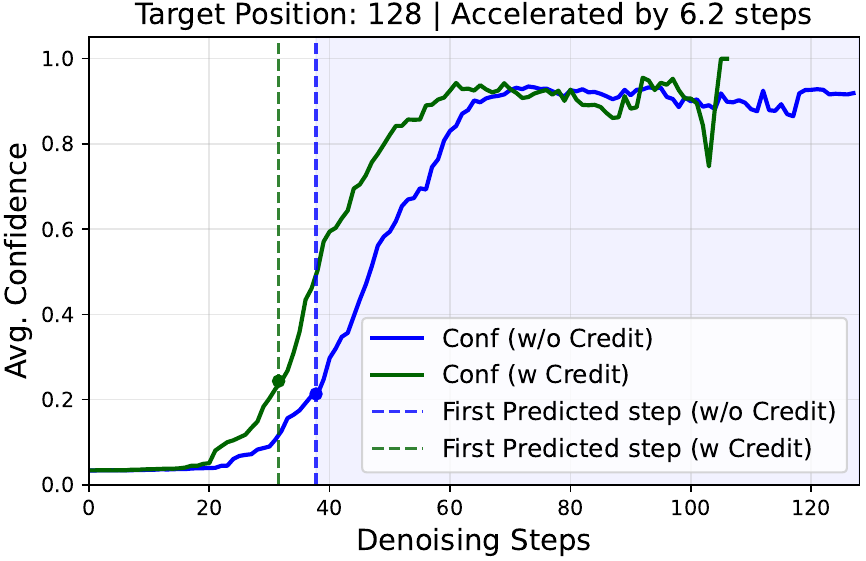}
    \caption{
        Impact of CreditDecoding on $x_0^{128,v}$ confidence. The shaded region marks steps where $x_0^{128,v}$ achieves top-$1$ confidence. CreditDecoding reduces steps by boosting confidence and entering this region earlier.
    }
    \label{fig:computational_redundancy}
    \vspace{-10pt}
\end{figure}

Figure~\ref{fig:computational_redundancy} (blue line) visualizes how the confidence of $x_0^{i,v}$ evolves over the denoising process. Many tokens are repeatedly predicted and remasked long before they are ultimately decoded. This is particularly pronounced under single-token decoding schemes, which, despite this inefficiency, often lead to higher final accuracy~\citep{feng2025theoreticalbenefitlimitationdiffusion}.

This observation reveals two limitations of existing methods:
\textbf{(i) Redundant computation.} Decisions are gated by instantaneous confidence, so correct hypotheses are repeatedly predicted and remasked until $s_t^{i}$ exceeds $\tau$.
\textbf{(ii) History-agnostic decoding.} Each step ignores past predictions; transient mispredictions can delay decoding and propagate errors to later denoising steps.

A natural idea is to \emph{promote early decoding} by boosting the confidence of target tokens so that they cross the threshold earlier.
For practicality and analytical tractability in the probability domain, we add a gain of the form $\log X$ ($X\ge 1$) to the target-token logit.
Specifically, for a target token $x_0^{i,v}$, we enhance its logit $l_t^{i,v}$ as:
\begin{equation}
\hat{l}_t^{i,v} = l_t^{i,v} + \log X,
\end{equation}
where $\hat{l}_t^{i,v}$ is the corresponding fused logit.

Let $p_t^{i,v}=p_\theta^{i}(v\mid x_t)$ denote the current probability of token $v$ at position $i$.
When $x_0^{i,v}$ is the predicted token at step $t$, we have $s_t^{i}=p_t^{i,v}$.
It is straightforward to show that the \emph{minimum} gain required to ensure $\hat{s}_t^{i}=\hat{p}_t^{i,v}\ge \tau$ is:
\begin{equation}
X \;\ge\; X_{\min}(p_t^{i,v},\tau)
=\frac{\tau}{1-\tau}\cdot(\frac{1}{p_t^{i,v}}-1).
\label{eq:xmin}
\end{equation}
Eq.~\ref{eq:xmin} shows that the required gain is highly sensitive to $\hat{p}_t^{i,v}$. However, early-step probabilities are unstable and the true target token is unknown. Naively applying $X_{\min}$ can amplify noise and lead to irreversible decoding errors.
This motivates a token-level \textit{\textbf{trace credit}} $C_t^{i,v}$ that aggregates historical evidence and serves as an adaptive gain for the current prediction.
Intuitively, credit measures a candidate's likelihood of converging to high confidence, enabling earlier yet safer decoding. The derivation is provided in Appendix~\ref{AppendixC1:appendix_xmin}.

\begin{figure}[t]
    \centering
    \includegraphics[width=\linewidth]{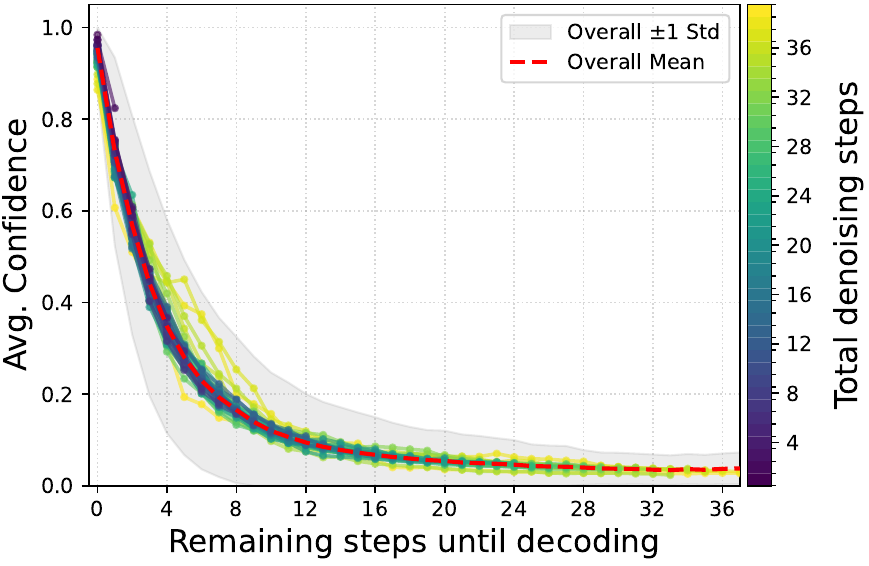}
    \caption{
    Temporal consistency. Results are obtained on GSM8K with threshold-based parallel decoding. Curves correspond to tokens grouped by total denoising steps.
    }
    \label{fig:conf_convergence}
    \vspace{-10pt}
\end{figure}
\begin{figure*}[t]
    \centering
    \includegraphics[width=0.32\linewidth]{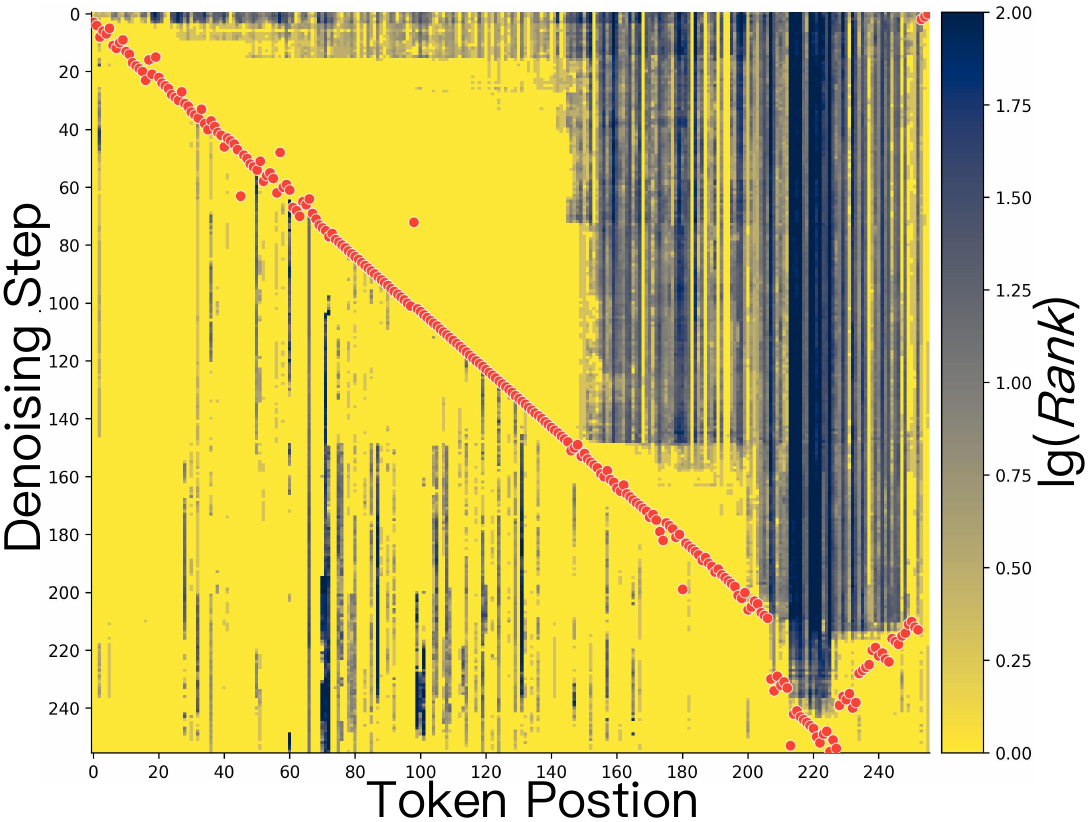}
    \includegraphics[width=0.32\linewidth]{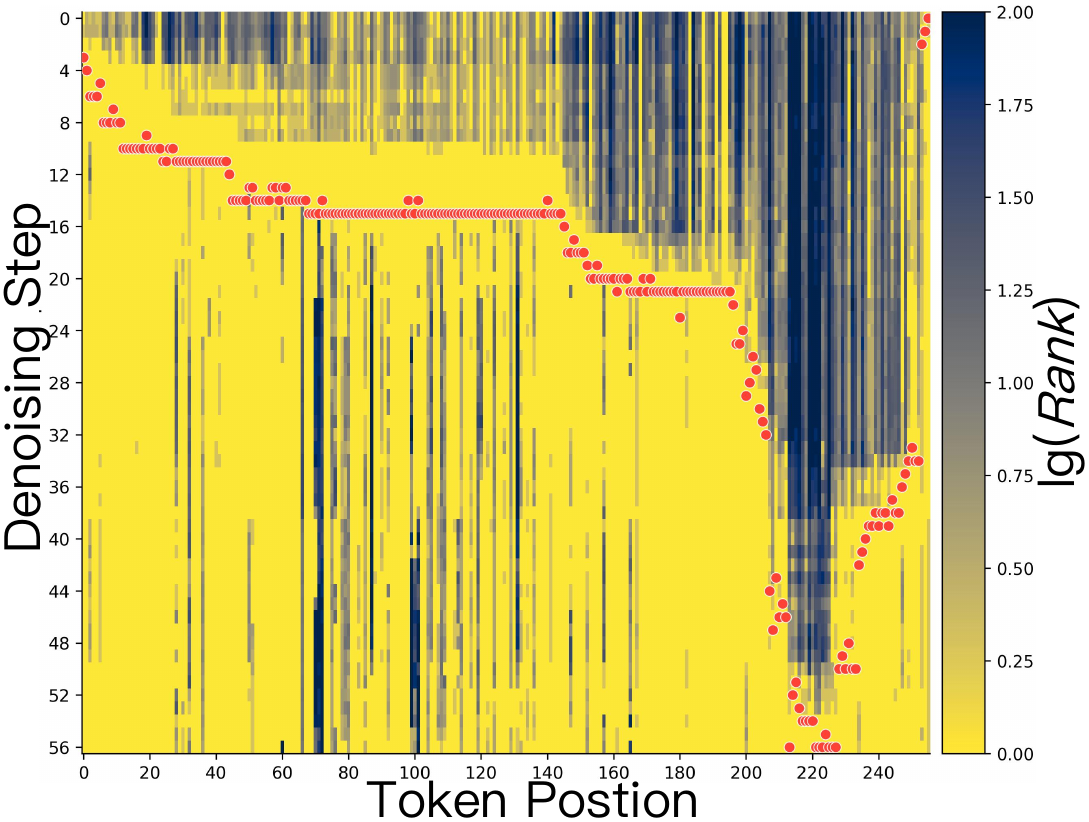}
    \includegraphics[width=0.32\linewidth]{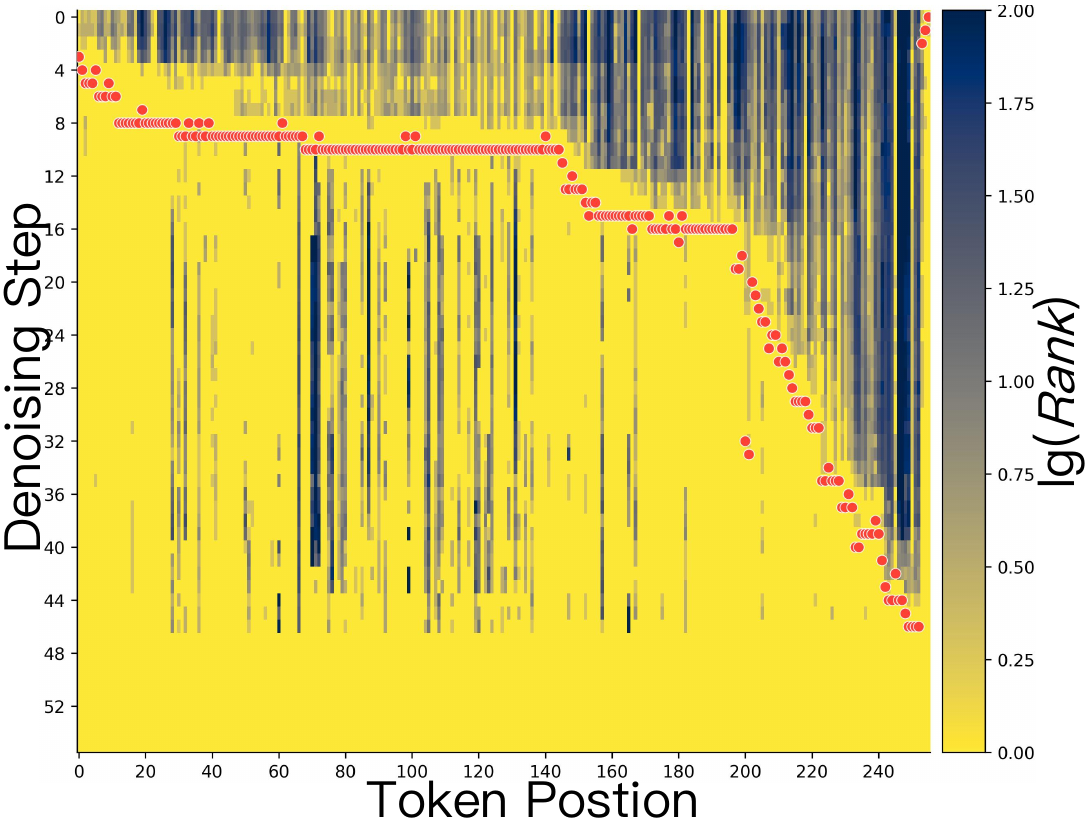}
    \caption{Confidence rank of the target token $x_0^{i,v}$(defined in~\ref{definition-of-target-token}) at each position $i$ during the denoising steps $s$ (\textbf{Left}: LLaDA, \textbf{Middle}: Fast-dLLM, \textbf{Right}: CreditDecoding). The red dots indicate that the model decodes tokens at position $i$ at the corresponding step $s$, which is also the \textit{Decoding Boundary} in Figure~\ref{AppendixC3:Ideal Decoding Boundary}.}
    \label{fig5:rank_correct}
\end{figure*}
\subsection{CreditDecoding}
\label{sec:3.2}
In this section, we introduce \emph{CreditDecoding}, a training-free mechanism that enables earlier and safer parallel decoding.

\textbf{Temporal Consistency }
Formally, we define the denoising trace $\mathcal{T}$ as the ordered collection of predicted tokens and their confidence during the denoising process.
Specifically, at each step $t$ from $T$ down to $0$, all predicted tokens $\tilde{x}_t$ and confidence scores $s_t$ are appended to $\mathcal{T}$.

As shown in Figure~\ref{fig:conf_convergence}, the confidence of the eventual target token $x_0^{i,v}$ exhibits a consistently increasing trend and quickly approaches 
$1$ as it nears decoding.
Moreover, this trend is largely insensitive to the length of $\mathcal{T}$, suggesting that the confidence of a target token is determined mainly by its \emph{denoising stage} (i.e., the remaining steps).

Motivated by this property, we use an EMA-based credit to aggregate temporal evidence into a stable prior, predicting whether the current greedy token $\tilde{x}_t^{i}$ will eventually be decoded.

\textbf{Definition of Trace Credit }
During inference, we maintain token-level credits independently for each position.
At each denoising step $t$, for every masked position $i\in M_t$ and token $v\in\mathcal{V}$, we maintain a credit value $C_t^{i,v}\in\mathbb{R}_{\ge 0}$ that accumulates historical evidence from 
$\mathcal{T}$
for supporting token $v$ at position $i$.
Given the predicted token $\tilde{x}_t^{i}$, credits are updated via an EMA-style rule that reinforces $\tilde{x}_t^{i}$:
\begin{equation}
\label{eq:credit-update}
C_t^{i,v} =
\begin{cases}
\beta \, C_{t+1}^{i,v} + \big(p_t^{i,v}\big)^{\gamma}, & v = \tilde{x}_t^{i}, \\[4pt]
\beta \, C_{t+1}^{i,v}, & \text{otherwise}.
\end{cases}
\end{equation}
Here $\beta\in(0,1)$ controls exponential decay, and $\gamma\in(0,1)$ is a concave transform that up-weights low-confidence values, which is fixed in our implementation.

As illustrated in Figure~\ref{fig:diagram}, this update rule balances two dynamics.
\textbf{(i) Global Decay}: $\beta$ gradually forgets stale evidence, especially from tokens not currently predicted, suppressing early-stage confidence fluctuations.
\textbf{(ii) Focused Enhancement}: for each position $i$, only the currently predicted token $\tilde{x}_t^{i}$ receives an additional credit boost, so credit accumulates mainly on tokens that consistently rank top-$1$ in confidence along the denoising trace rather than on transient spikes.

For completeness, we also explore a variant that aggregates historical evidence for all tokens. Details are given in Appendix~\ref{Appendix:fullcredit}.
Ablations in Appendix~\ref{Appendix:focus_robustness} confirm that the focused enhancement strategy ensures the optimal speed-accuracy trade-off, outperforming \textit{Top-$K$ Enhancement} variants.

Crucially, we fuse credit with logits to obtain a sharpened predictive distribution:
\begin{equation}
\label{eq:logits-enhance}
\hat{l}_t^{i,v} = l_t^{i,v} + \alpha \cdot \log\!\big(C_t^{i,v} + 1\big)
\end{equation}
where $\alpha>0$ controls the strength of the credit-based prior.
Eq.~\ref{eq:logits-enhance} is equivalent to applying a multiplicative gain to $p_t^{i,v}$ in the probability domain, yielding $\hat{p}_t^{i,\cdot}=\mathrm{Softmax}\!\big(\hat{l}_t^{i}\big)$.

Accordingly, the confidence score becomes $\hat{s}_t^{i}=\max_{v\in\mathcal V}\hat{p}_t^{i,v}$, and the predicted token under the enhanced distribution is $\tilde{x}_t^{i}=\arg\max_v \hat{p}_t^{i,v}$.
Tokens predicted consistently across steps thus accumulate credit, inducing a growing effective gain that enables earlier decoding and reduces redundant remasking.
Conversely, unstable candidates are down-weighted by decay, improving robustness to transient fluctuations.
Importantly, CreditDecoding modifies only the output logits and seamlessly integrates with mainstream inference optimizations to achieve cumulative speedups. 

In practice, we apply CreditDecoding only within the \emph{current denoising block} for inference efficiency. Algorithm~\ref{alg:creditdecode} provides the complete procedure.

\subsection{Tuning-Free Schedule}
\label{tuning-free}
To avoid hyperparameter tuning, we propose a step-adaptive schedule that couples credit strength to the denoising stage.

Let $\eta_t$ denote the mask ratio at step $t$.
We set $\gamma=1$ and use step-dependent coefficients $\beta_t=\alpha_t=1-\eta_t$ to reflect the denoising progress, where $\alpha_t$ is the logit-fusion weight in Eq.~\ref{eq:logits-enhance}.

In early steps, it down-weights trace credit to counteract unreliable confidence.
As denoising progresses and $\eta_t$ decreases, predictions stabilize and credit strength increases automatically.

\begin{table*}[h]
\centering
\caption{
Main benchmark results \textbf{w/ Early Stop} across eight datasets on LLaDA-8B-Instruct and LLaDA-MoE-Instruct(Gen Length=256, Block Size=64). Cells show \textbf{Score} (top, relative to LLaDA) and \textbf{TPF} (bottom, improvement over Fast-dLLM). The last row (\textbf{Avg.}) reports the mean Score and mean TPF over the eight datasets.
}
\renewcommand{\arraystretch}{1.16}
\setlength{\tabcolsep}{3pt}
\footnotesize
\resizebox{0.89\textwidth}{!}{
\begin{tabular}{l ccc | ccc}
\toprule
\multirow{2}{*}{\centering Benchmark} 
&\multicolumn{3}{c}{\centering \textbf{LLaDA-8B-Instruct}} 
&\multicolumn{3}{c}{\centering \textbf{LLaDA-MoE-Instruct}} \\
& Baseline & Fast-dLLM & CreditDecoding  & Baseline & Fast-dLLM & CreditDecoding  \\
\midrule
\multirow{2}{*}{\centering MMLU
            ${}^{^{SCORE}}_{_{TPF}}$}
            & 62.46 
            & 62.43 $_{\textcolor{red!50!black}{-0.03}}$ 
            & \textbf{63.78 $_{\textcolor{green!50!black}{+1.32}}$}
            & 64.08 
            & 64.08 $_{0.00}$
            & \textbf{64.21 $_{\textcolor{green!50!black}{+0.13}}$}
            \\
            & 1 & 2.86
            & \textbf{4.57} (\textit{\textcolor{green!50!black}{+56\%}}) 
            & 1 
            & 2.16 
            & \textbf{2.46} (\textit{\textcolor{green!50!black}{+14\%}}) 
            \\
\midrule
\multirow{2}{*}{\centering SQuAD2.0
            ${}^{^{SCORE}}_{_{TPF}}$}
            & 91.43 
            & 91.43 $_{0.00}$
            & \textbf{91.71 $_{\textcolor{green!50!black}{+0.28}}$} 
            & 86.88 
            & 86.88 $_{0.00}$
            & \textbf{87.27 $_{\textcolor{green!50!black}{+0.39}}$ }
            \\
            & 1 & 13.55
            & \textbf{16.84} (\textit{\textcolor{green!50!black}{+24\%}})
            & 1 
            & 7.09 
            & \textbf{9.64} (\textit{\textcolor{green!50!black}{+36\%}}) 
            \\

\midrule
\multirow{2}{*}{\centering DROP
            ${}^{^{SCORE}}_{_{TPF}}$}
            & 82.86 
            & 82.74 $_{\textcolor{red!50!black}{-0.12}}$ 
            & \textbf{82.78 $_{\textcolor{red!50!black}{-0.08}}$} 
            & \textbf{80.16} 
            & \textbf{80.16} $_{0.00}$
            & 79.72 $_{\textcolor{red!50!black}{-0.44}}$ 
            \\
            & 1 & 2.93
            & \textbf{3.79} (\textit{\textcolor{green!50!black}{+29\%}})
            & 1 
            & 2.73 
            & \textbf{3.28} (\textit{\textcolor{green!50!black}{+20\%}}) 
            \\

\midrule
\multirow{2}{*}{\centering KorBench
            ${}^{^{SCORE}}_{_{TPF}}$}
            & 33.12 
            & 33.20 $_{\textcolor{green!50!black}{+0.08}}$ 
            & \textbf{35.04 $_{\textcolor{green!50!black}{+1.92}}$ }
            & 36.72 
            & \textbf{36.88} $_{\textcolor{green!50!black}{+0.16}}$ 
            & 36.48 $_{\textcolor{red!50!black}{-0.24}}$ 
            \\
            & 1 & 3.72
            & \textbf{5.03} (\textit{\textcolor{green!50!black}{+35\%}})
            & 1 
            & 2.36 
            & \textbf{3.28} (\textit{\textcolor{green!50!black}{+38\%}}) 
            \\

\midrule
\multirow{2}{*}{\centering HumanEval
            ${}^{^{SCORE}}_{_{TPF}}$}
            & 34.76 
            & 34.15 $_{\textcolor{red!50!black}{-0.61}}$ 
            & \textbf{36.59 $_{\textcolor{green!50!black}{+1.83}}$} 
            & \textbf{51.22}
            & \textbf{51.22 $_{0.00}$}
            & \textbf{51.22 $_{0.00}$}
            \\
            & 1 & 3.82 
            & \textbf{4.69} (\textit{\textcolor{green!50!black}{+23\%}}) 
            & 1 
            & 4.97 
            & \textbf{6.00} (\textit{\textcolor{green!50!black}{+21\%}}) 

            \\

\midrule
\multirow{2}{*}{\centering LCB 
            ${}^{^{SCORE}}_{_{TPF}}$}
            & \textbf{8.15} 
            & \textbf{8.15} $_{0.00}$
            & 7.54 $_{\textcolor{red!50!black}{-0.61}}$ 
            & 13.88 
            & 14.04 $_{\textcolor{green!50!black}{+0.16}}$ 
            & \textbf{14.37 $_{\textcolor{green!50!black}{+0.49}}$ }
            \\
            & 1 & 1.93
            & \textbf{2.17} (\textit{\textcolor{green!50!black}{+12\%}})
            & 1 
            & 2.43 
            & \textbf{2.81} (\textit{\textcolor{green!50!black}{+16\%}}) 
            \\
\midrule
\multirow{2}{*}{\centering GSM8K 
            ${}^{^{SCORE}}_{_{TPF}}$}
            & 77.94 
            & \textbf{78.47 $_{\textcolor{green!50!black}{+0.53}}$ }
            & 77.18 $_{\textcolor{red!50!black}{-0.76}}$ 
            & 74.37 
            & 74.45 $_{\textcolor{green!50!black}{+0.08}}$ 
            & \textbf{74.98 $_{\textcolor{green!50!black}{+0.61}}$} 
            \\
            & 1 
            & 3.22 
            & \textbf{3.87} (\textit{\textcolor{green!50!black}{+20\%}})
            & 1 
            & 2.28 
            & \textbf{2.68} (\textit{\textcolor{green!50!black}{+18\%}}) 
            \\
\midrule
\multirow{2}{*}{\centering MATH 
            ${}^{^{SCORE}}_{_{TPF}}$}
            & 37.30 
            & 37.04 $_{\textcolor{red!50!black}{-0.26}}$ 
            & \textbf{37.24 $_{\textcolor{red!50!black}{-0.06}}$ }
            & 36.02 
            & 35.84 $_{\textcolor{red!50!black}{-0.18}}$ 
            & \textbf{36.28 $_{\textcolor{green!50!black}{+0.26}}$} 
            \\
            & 1
            & 2.42 
            & \textbf{2.84} (\textit{\textcolor{green!50!black}{+17\%}})
            & 1 
            & 2.35 
            & \textbf{2.71} (\textit{\textcolor{green!50!black}{+15\%}}) 
            \\
\midrule
\multirow{2}{*}{\centering \textbf{Average} 
            ${}^{^{SCORE}}_{_{TPF}}$}
            & 53.50 
            & 53.45$_{\textcolor{red!50!black}{-0.05}}$
            & \textbf{53.98$_{\textcolor{green!50!black}{+0.48}}$ }
            & 55.42 
            & 55.44 $_{\textcolor{green!50!black}{+0.02}}$ 
            & \textbf{55.57 $_{\textcolor{green!50!black}{+0.15}}$ }
            \\
            & 1 
            & 4.31 
            & \textbf{5.48} (\textit{\textcolor{green!50!black}{+27\%}})
            & 1 
            & 3.30 
            & \textbf{4.11} (\textit{\textcolor{green!50!black}{+25\%}}) 
            \\

\bottomrule
\label{tab:Main-Results}
\end{tabular}
}
\vspace{-10pt}
\end{table*}

\section{Experiments and Analyses}

\subsection{Experimental Setup}

\textbf{Implementation } We implement CreditDecoding on LLaDA-8B-Instruct~\citep{nie2025large} and LLaDA-MoE-Instruct~\citep{llada-moe-7b-a1b-base}. Inference settings differ across experiments and may deviate slightly from the original papers. We detail them in the corresponding sections. All experiments are conducted on NVIDIA H20-3e 140 GB GPUs.

In the main experiments, we set the generation length and number of steps to 256. Additional experiments with different generation lengths are reported in Section~\ref{sec:5.5}. Based on the analyses in Appendix~\ref{AppendixC4:block_length} and Section~\ref{sec:5.3}, we set the block size to 64, and the hyperparameters to $\alpha=0.65$, $\beta =0.7$ and $\gamma=0.2$, \emph{which are tuned on LLaDA-8B-Instruct}. All experiments are conducted~\textit{w/ Early Stop} for meaningful TPF and ~\textit{ w/o KV cache}, unless  specified in Section~\ref{sec:5.6}

\textbf{Evaluation Tasks }
In the main experiments, we comprehensively evaluate CreditDecoding on eight datasets spanning five categories. Specifically, we evaluate inference performance on DROP \citep{dua2019dropreadingcomprehensionbenchmark} and KorBench \citep{ma2025korbenchbenchmarkinglanguagemodels}, language understanding on SQuAD2.0, knowledge assessment on MMLU \citep{hendrycks2021measuringmassivemultitasklanguage}, coding ability on OpenAI HumanEval \citep{chen2021evaluatinglargelanguagemodels} and LiveCodeBench \citep{jain2024livecodebenchholisticcontaminationfree}, and mathematical reasoning on GSM8K  \citep{cobbe2021trainingverifierssolvemath} and MATH \citep{hendrycks2021measuringmathematicalproblemsolving}.
In the ablation, scaling, and other analysis experiments, due to computational constraints, we select five representative datasets across categories: MMLU, SQuAD2.0, KorBench, HumanEval, and GSM8K.

\textbf{Evaluation Metrics }
We adopt the standard performance metrics for each evaluation dataset, as detailed in Appendix~\ref{AppendixB:Eval Cfg}. In addition, to examine whether CreditDecoding can mitigate redundant computation inherent in traditional dLLMs, we utilize \emph{TPF} (Tokens Per Forward) to evaluate dLLM inference efficiency. 
We report single-run results for each setting.

\textbf{Early Stop } In dLLM, early stopping changes the generated length and thus significantly affects TPF. Some studies disable it to boost TPF, as the model quickly outputs the EOS token. However, in practical applications it is usually enabled to avoid redundant outputs. 
We \emph{enable} Early Stop by default, except in Figure~\ref{fig:Orthogonality} for better comparison.

\subsection{Main Results }

As shown in Table~\ref{tab:Main-Results}, we evaluate our method on eight datasets using LLaDA-8B-Instruct and LLaDA-MoE-Instruct. We use each benchmark’s default performance metric and report TPF for inference speed, with TPF of the baseline normalized to 1. TPF of CreditDecoding thus directly reflects speedup relative to the baseline.

Overall, \textbf{\textit{CreditDecoding outperforms the baseline and the SOTA Fast-dLLM in both performance and speed}}. It achieves a 5.48$\times$ speedup and 0.48 performance gain on LLaDA-8B-Instruct, and a 4.10$\times$ speedup on LLaDA-MoE-Instruct. 
Furthermore, the increase in TPF translates directly into significant \textbf{\textit{end-to-end speedups}}. As detailed in Appendix~\ref{AppendixC7:Orthogonality} and~\ref{Appendix:end_to_end_efficiency}, CreditDecoding consistently improves Tokens Per Second (TPS) across different  acceleration scenarios on both LLaDA-8B-Instruct and LLaDA-MoE-Instruct.
\begin{table*}[t]
    \centering
    \caption{Performance comparison across different model paradigms and scales. Results are reported as the average \textbf{Score / TPF} over the evaluated datasets with default hyperparameters. CD refers to CreditDecoding.}
    \label{tab:Generalizability}
    \setlength{\tabcolsep}{4pt} 
    \renewcommand{\arraystretch}{0.92}
    
    \resizebox{0.85\textwidth}{!}{
    \begin{tabular}{llll ccc}
        \toprule
        \textbf{Paradigm} & \textbf{Arch.} & \textbf{Model} & \textbf{Params} & \textbf{Fast-dLLM} & \textbf{CD} & \textbf{CD-Adaptive} \\
        \midrule
        \multirow{2}{*}{\shortstack[l]{Pure\\Diffusion}} & Dense & LLaDA-8B-Ins & 8B & \underline{60.38} / 5.49 & \textbf{60.86} / \underline{7.00} & 60.23 / \textbf{7.74} \\
         & MoE & LLaDA-MoE-Ins & 7B-A1B & 62.70 / 3.77 & \underline{62.83} / \underline{4.81} & \textbf{63.74} / \textbf{5.27} \\
        
        \midrule
        \multirow{3}{*}{\shortstack[l]{Block\\Diffusion}} 
         & Dense & SDAR-Chat-b32 & 8B & 65.37 / 1.72 & \underline{65.51} / \textbf{1.91} & \textbf{66.37} / \underline{1.83} \\
        & MoE & LLaDA2-Mini & 16B-A1B & 76.23 / 2.01 & \underline{76.69} / \textbf{2.26} & \textbf{76.92} / \underline{2.15} \\
         & MoE & LLaDA2-Flash & 100B-A6B & 84.89 / 2.48 & \underline{84.53} / \textbf{2.79} & \textbf{84.97} / \underline{2.68} \\

        \bottomrule
        
    \end{tabular}
    }
    \vspace{-5pt}
\end{table*}

Figure~\ref{fig5:rank_correct} illustrates that after applying CreditDecoding, the red line representing the decoding boundary becomes more horizontal, indicating more tokens decoded per step and higher efficiency. While the baseline requires all 256 steps, CreditDecoding completes decoding in 50 steps, consistent with the ~5$\times$ speedup in Table~\ref{tab:Main-Results}. Yellow and blue denote high and low confidence respectively, with their boundary marking the step where the model predicts the target token for the first time. The gap between the red line and the boundary shows the redundant steps introduced by the decoding method. Traditional dLLMs discard remasked token information, necessitating repeated predictions and creating a gap between the red line and the yellow-blue boundary.

\textbf{\textit{Two observations in Figure~\ref{fig5:rank_correct} explain this acceleration:}} \textbf{(i)} The yellow-blue boundary moves upward, as trace credit acts as a temporal~\emph{low-pass filter} that suppresses confidence spikes on incorrect tokens while reinforcing those with a steadily increasing trend (Figure~\ref{fig:conf_convergence}), thereby predicting the target token earlier and improving the upper limit of the red line; \textbf{(ii)} CreditDecoding decreases the gap between the red line and the yellow-blue boundary, because earlier decoding of correct tokens improves the intermediate context for subsequent steps, creating a positive feedback loop (Appendix~\ref{AppendixC5:Contextual Cumulative Effect}) that mitigates redundant steps.

\subsection{Hyperparameter Ablation Study}
\label{sec:5.3}

As discussed in Section~\ref{sec:3.2}, CreditDecoding has three hyperparameters: $\beta$ for global decay, $\alpha$ for logits fusion, and $\gamma$ for concave amplification. We fix $\gamma$ and perform ablation studies on $\alpha$ and $\beta$ in the range [0, 0.95] with a step size of 0.05.

\begin{figure}[t]
    \centering
    \includegraphics[width=\linewidth]{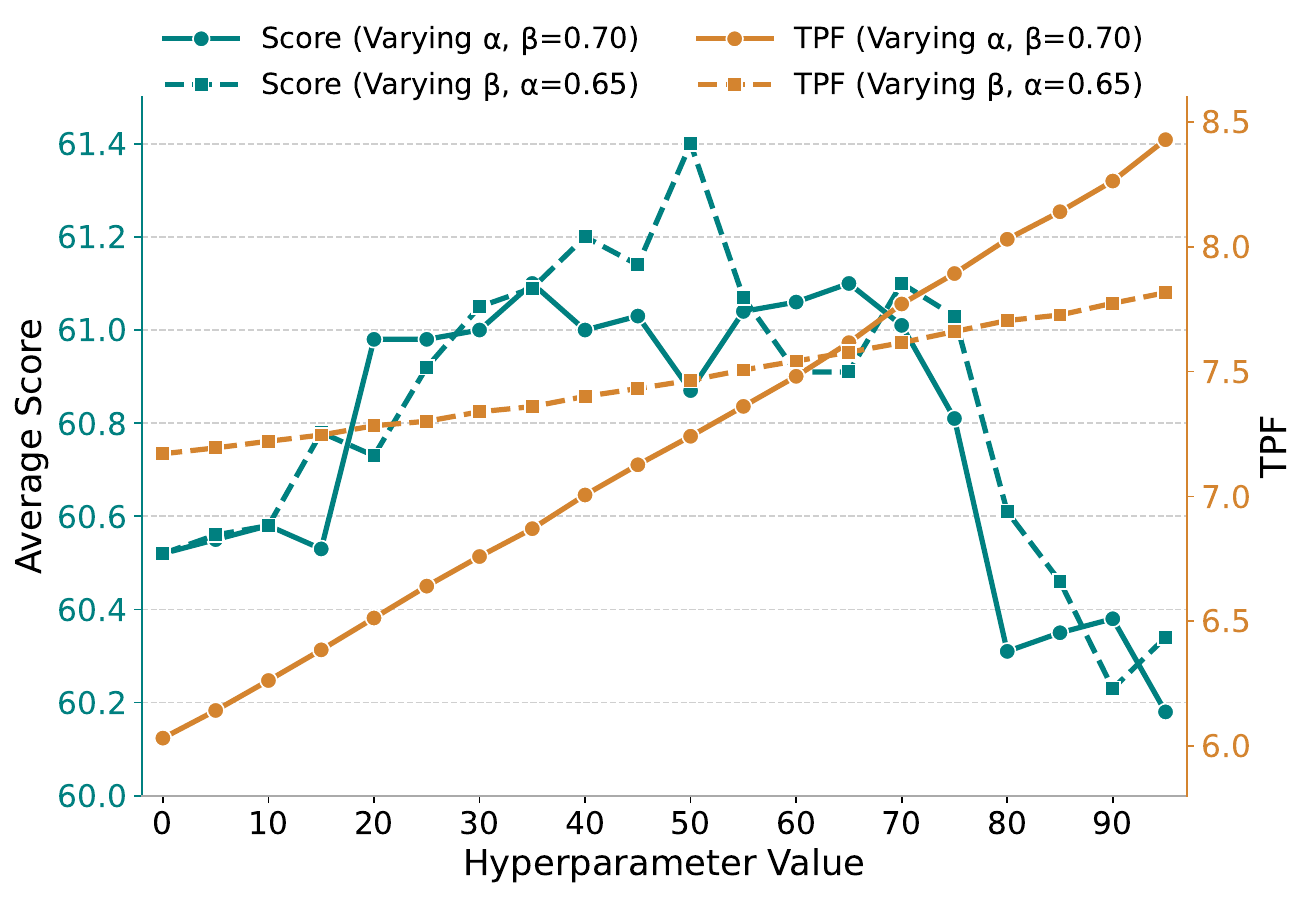}
    \caption{Hyperparameter Ablation 1. Green: average score; Yellow: average TPF. \textbf{Solid}: varying $\alpha$ with fixed $\beta=0.70$; \textbf{Dashed}: varying $\beta$ with fixed $\alpha=0.65$.}
    \label{fig:hyper_param}
    \vspace{-10pt}
\end{figure}

Figure~\ref{fig:hyper_param} shows that performance fluctuates slightly with $\alpha$ and $\beta$, peaking around $\beta=0.5$ and $\beta=0.7$, while $\alpha$ remains stable within [0.2, 0.65]. \textbf{\textit{Larger values of both parameters increase TPF by accumulating more trace credit, but may reduce accuracy.}} We select $\alpha=0.65$ and $\beta=0.7$ as they provide a good trade-off, yielding strong performance and high TPF across five datasets, though optimal values vary by dataset.

\begin{figure}[t]
    \centering
    \includegraphics[width=\linewidth]{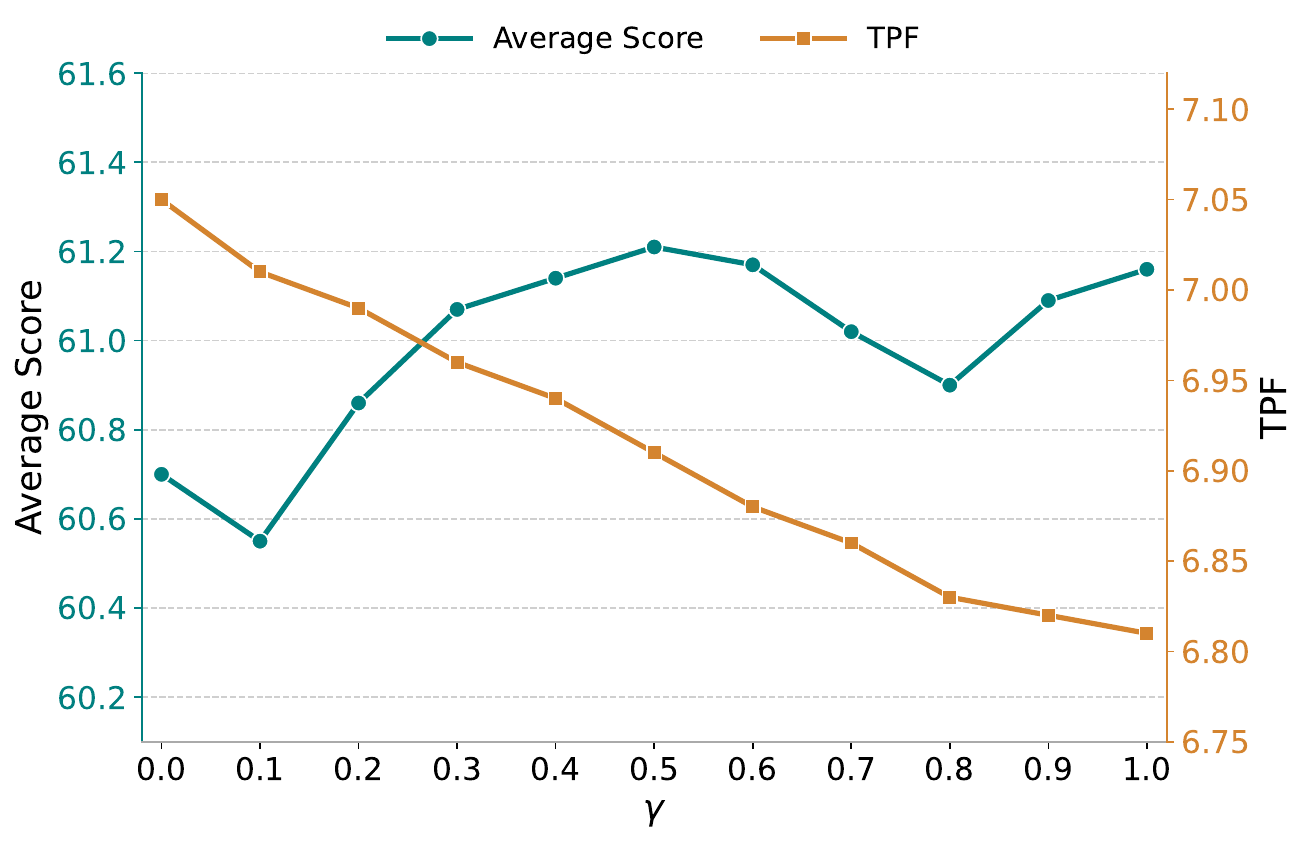}
    \caption{Hyperparameter Ablation 2. Green: average  score; Yellow: average TPF. \textbf{Solid}: varying $\gamma$ with fixed $\beta=0.70$ and $\alpha=0.65$.}
    \label{fig:hyper_param_2}
    \vspace{-10pt}
\end{figure}

Furthermore, we investigate the impact of $\gamma$. As detailed in Figure~\ref{fig:hyper_param_2}, TPF exhibits a strict negative correlation with $\gamma$. Meanwhile, similar to the trend observed in Figure~\ref{fig:hyper_param}, the generation score follows a non-linear trajectory, peaking at $\gamma=0.5$ before showing a slight recovery as $\gamma$ approaches $1.0$. Consequently, we decouple $\gamma$ by fixing it at a relatively small value of $0.2$ in our main evaluations to ensure a robust acceleration effect.


\subsection{Generalizability}
\label{sec:5.4}

In this section, we demonstrate the generalizability of CreditDecoding by conducting experiments across models characterized by diverse paradigms and scales. As shown in Table \ref{tab:Generalizability}, in addition to the LLaDA-8B-Ins and LLaDA-MoE-Ins models discussed in the main results, we extend our evaluation to include SDAR-8B-Chat-b32~\citep{cheng2025sdarsynergisticdiffusionautoregressionparadigm}, LLaDA2-Mini~\citep{bie2025llada2} and LLaDA2-Flash~\citep{bie2025llada2}. Collectively, these experiments encompass various training paradigms and architectures, spanning parameter scales from 8B to 100B. We employ dInfer \citep{ma2025dinfer} for inference, configuring the generation length to 2048 and the block size to 32. Due to computational constraints, we limit our comparison to Fast-dLLM, CreditDecoding, and the CreditDecoding-Adaptive, which adapts to the step mask ratio introduced in Section~\ref{tuning-free}, omitting weaker baselines that are incompatible with parallel decoding.

Results in Table~\ref{tab:Generalizability} yield two critical insights. First, \textbf{\textit{CreditDecoding strategy delivers more substantial efficiency gains in Pure Diffusion training paradigms}} due to its larger attention context window. Second, and most significantly, \textbf{\textit{CreditDecoding-Adaptive serves as a universal, tuning-free solution}}. Addressing the dataset-dependent nature of denoising traces $\mathcal{T}$ (Appendix~\ref{AppendixC6:dataset-dependent}), it eliminates the need for task-specific hyperparameter tuning, thereby reducing deployment overhead. It yields superior generation scores and remains significantly faster than Fast-dLLM, balancing performance and usability with minimal efficiency cost.

\subsection{Scalability}
\label{sec:5.5}

Current research on dLLMs typically evaluates generation lengths of 128 or 256, with a few studies extending to 512. However, the primary advantage of dLLMs over autoregressive models—parallel decoding—is largely underutilized at such short lengths. To address this, in this section we scale the generation length to 1024 and 4096, aiming to provide insights into the potential of dLLMs for long-text generation.

\begin{figure}[t]
    \centering
    \includegraphics[width=\linewidth]{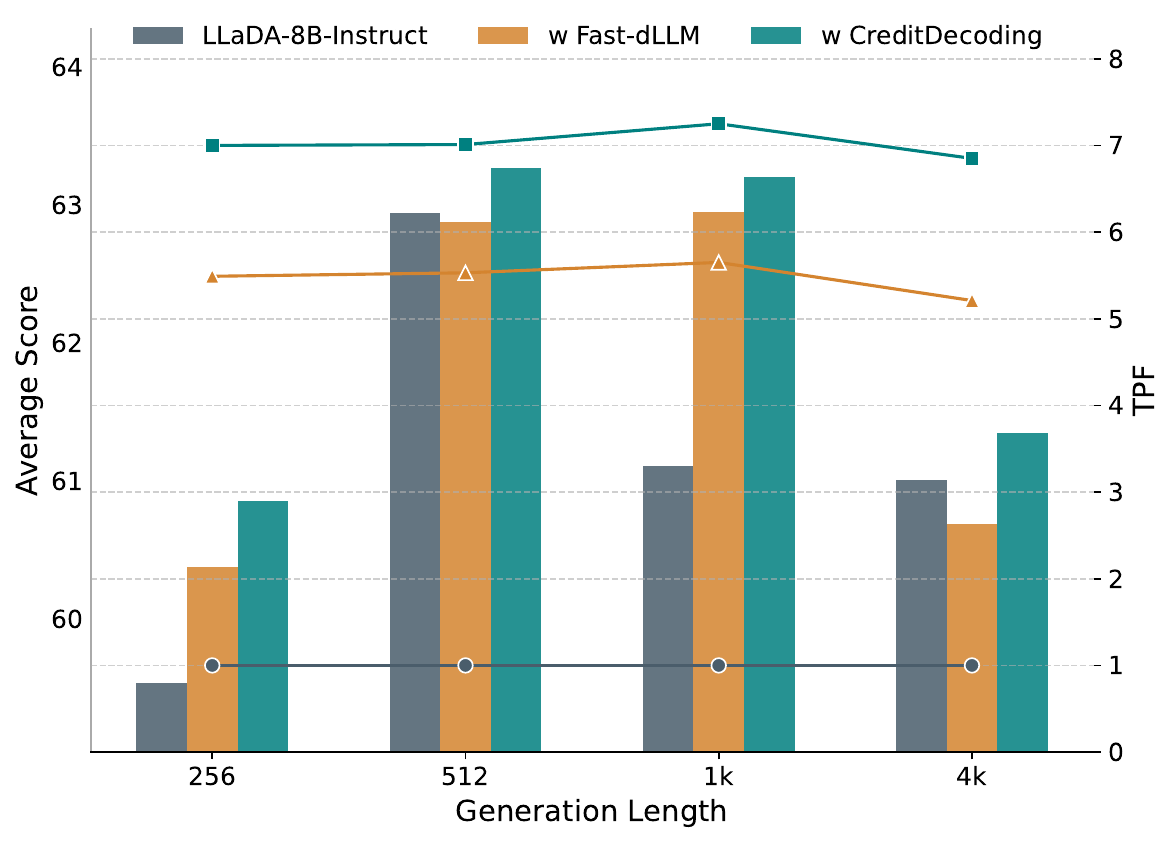}
    \caption{Scalability of CreditDecoding on LLaDA. Lines indicate average TPF; bars show average score.}
    \label{fig:Scalability}
    \vspace{-10pt}
\end{figure}

We fix the number of steps equal to the generation length and set the block size to 64. Figure~\ref{fig:Scalability} shows that both the baseline and CreditDecoding peak at length 512 and gradually degrade as length increases. Notably, \textbf{\textit{CreditDecoding consistently outperforms the baseline and scales better with generation length}}, demonstrating stronger robustness in long-context scenarios.

Crucially, the method sustains a consistent ~7$\times$ TPF speedup across settings. As text length grows, \textbf{\textit{this relative speedup translates into increasingly substantial absolute latency reductions}}, highlighting its practical value in long-context scenarios.

\subsection{Orthogonality}
\label{sec:5.6}
\begin{table*}[t]
    \centering
    \caption{Performance and efficiency comparison of CreditDecoding across different decoding strategies.}
    \label{tab:decoding_comparison}
    \setlength{\tabcolsep}{4pt}
    \renewcommand{\arraystretch}{0.85}
    \begin{tabular}{lcccc}
    \toprule
    \textbf{Decoding Algorithm} & \textbf{Score} & \textbf{Score (w/ Credit)} & \textbf{TPF} & \textbf{TPF (w/ Credit)} \\
    \midrule
    Top-2            & 57.59 & \textbf{57.73} (\textit{\textcolor{green!50!black}{+0.14}}) & 2.00 & 2.00 (\textit{\textcolor{gray!80}{+0.00}})\\
    Top-4            & 52.92 & \textbf{53.79} (\textit{\textcolor{green!50!black}{+0.87}}) & 4.00 & 4.00 (\textit{\textcolor{gray!80}{+0.00}})\\
    Top-8            & 46.94 & \textbf{47.18} (\textit{\textcolor{green!50!black}{+0.24}}) & 8.00 & 8.00 (\textit{\textcolor{gray!80}{+0.00}})\\
    Top-$K$ Margin   & 60.44 & \textbf{61.20} (\textit{\textcolor{green!50!black}{+0.76}}) & 5.49 & \textbf{6.27} (\textit{\textcolor{green!50!black}{+0.78}}) \\
    Threshold (0.95) & 60.38 & \textbf{60.86} (\textit{\textcolor{green!50!black}{+0.48}}) & 5.49 & \textbf{7.00} (\textit{\textcolor{green!50!black}{+1.51}}) \\
    \bottomrule
    \end{tabular}
    \vspace{-5pt}
\end{table*}
CreditDecoding operates purely on the model logits and does not interfere with the \emph{sample} or \emph{select} procedures. 
This design makes it a \textbf{\textit{plug-and-play}} post-processing module that is naturally orthogonal to both (i) system-level inference optimizations such as compiler-level acceleration \citep{pytorch20}, and (ii) algorithmic acceleration strategies for dLLMs such as threshold decoding, KV cache variants, and EOS early stopping. 

In all cases, we keep the hyperparameters and selection rules of the baseline methods unchanged, and only apply CreditDecoding on top. As demonstrated in Figure~\ref{fig:Orthogonality}, our experiments confirm this orthogonality. 
When CreditDecoding is combined with these methods, we observe that the speedup is preserved while the performance drop is partially mitigated, showing that historical trace credit acts as a stabilizing prior under aggressive compression. 

For algorithmic acceleration, threshold parallel decoding (Fast-dLLM w/o KV cache ~\citep{wu2025fastdllmtrainingfreeaccelerationdiffusion}) typically accelerates but at the cost of lower accuracy.
For EOS early stopping, CreditDecoding complements the strategy by improving robustness of token selection in earlier steps. Adding CreditDecoding further boosts both speed and accuracy, with $4.7\times$ acceleration over the baseline with an additional $+0.63$ improvement in average score. 
Similarly, CreditDecoding alleviates the accuracy loss of threshold decoding and complements KV cache methods by reducing redundant iterations within cached segments.
These results demonstrate that \textbf{\textit{CreditDecoding can be seamlessly integrated with existing optimizations to deliver consistent performance gains}}.

Appendix~\ref{AppendixC7:Orthogonality} provides details of our acceleration methods and orthogonality experimental results for CreditDecoding on LLaDA and LLaDA-MoE, including \textit{\textbf{inference throughput (TPS)}}.

\subsection{Versatility}
\label{sec:5.7}

To comprehensively evaluate CreditDecoding across diverse mainstream decoding schemes, we supplement our study with \textit{Top-$K$} decoding experiments ($K \in \{2, 4, 8\}$), where a strictly fixed number of highest-confidence tokens are decoded at each step. 

It should be noted that CreditDecoding primarily accelerates decoding by accumulating historical trajectories into the trace credit, enabling high-confidence tokens to reach the decoding threshold earlier. This acceleration is directly reflected in the improvement of TPF. However, Top-$K$ decoding intrinsically fixes the token count per step, which inherently locks the TPF and overall decoding speed. Consequently, the TPF remains identical across all decoding methods under this setting, structurally neutralizing, the inference speedup that acceleration techniques would otherwise provide. Furthermore, consistent with prior studies, fixing the token count per step often leads to substantial performance degradation compared to threshold-based methods, which typically offer superior performance and lower latency. This phenomenon is also clearly reflected in our empirical results.

To demonstrate the robustness of CreditDecoding across different selection rules, we additionally incorporate the \textit{Top-$K$ margin} decoding method. The comparative results are detailed in Table~\ref{tab:decoding_comparison}.

As illustrated above, while the fixed decoding rate of Top-$K$ fundamentally nullifies the potential for acceleration (TPF remains constant), CreditDecoding still yields consistent generation performance improvements. This demonstrates that leveraging the historical information stored in the trace credit enhances prediction accuracy even when speedups are bottlenecked. More importantly, under flexible decoding settings such as Top-$K$ margin and threshold-based decoding, the integration of CreditDecoding consistently delivers substantial enhancements in both generation quality and inference efficiency.
\section{Conclusion}

The history-agnostic decoding process of diffusion language models inherently suffers from substantial computational redundancy. To address this, we propose CreditDecoding, a training-free decoding strategy orthogonal to mainstream acceleration methods. Specifically, it assigns a trace credit to each candidate token by accumulating historical logit predictions from remasked steps, and subsequently fuses this credit into the current logits.

By fully leveraging the model’s past predictions on remasked tokens, CreditDecoding reduces the complexity of current inference steps. This leads to simultaneous improvements in performance and decoding efficiency, achieving a 5.48$\times$ speedup and an average accuracy gain of $+$0.48 compared to standard LLaDA-8B-Instruct.

Ultimately, CreditDecoding aims to approach the theoretical limit of acceleration (indicated by the yellow-blue boundary in Figure~\ref{fig5:rank_correct}). As more powerful base models emerge, the acceleration ceiling of CreditDecoding will rise, leading to further efficiency gains in model inference.

\section*{Limitations}

Although CreditDecoding provides consistent acceleration and accuracy gains, it also has several limitations.

First, it relies on the accumulation of historical evidence along the denoising trace, and thus requires a certain number of denoising steps to build effective trace credit.
For pure-diffusion dLLMs that adopt small block sizes to trade inference speed for better performance, and for dLLMs that are explicitly trained for aggressive parallel decoding whose logit distributions are sharp with fast confidence convergence, our approach may not have enough denoising steps to accumulate sufficient trace credit.
In these cases, the acceleration benefit becomes limited.
Experiments in Appendix~\ref{AppendixC4:block_length} also show that increasing the block size generally allows stronger credit accumulation and leads to larger speedup.

Second, due to variations in dataset characteristics, such as data pattern or sample difficulty, and inherent model behavior, the optimal setting for $\alpha$, $\beta$ and $\gamma$ may differ across tasks and models.
Although we propose a tuning-free adaptive variant in Sec.~\ref{tuning-free} that yields stable and consistent improvements across model architectures, benchmarks, and parameter scales, it does not always achieve the theoretically optimal performance–efficiency trade-off.
In practice, obtaining the best benefit still requires lightweight hyperparameter search on a small validation set.

Third, we observe a slight performance degradation on reasoning-intensive tasks, such as mathematical reasoning and complex code generation. These tasks inherently possess strong causal dependencies and unstable confidence convergence, which can interfere with the trace credit accumulation during highly parallel decoding. A detailed failure mode analysis regarding these tasks is provided in Appendix~\ref{Appendix:failure_mode}. 
\section*{Acknowledgements} 
We thank the anonymous reviewers and the area
chair for their constructive comments. This work was supported by Ant Group Research Fund, and supported in part by the National Natural Science Foundation of China (Grant No. 62595733, 62561160155, 62325109), the Shanghai 'The Belt and Road' Young Scholar Exchange Grant (Grant No. 24510742000), and the Key R\&D Program of Zhejiang (Grant No. 2025C01104).
\bibliography{CreditDecoding}

@misc{ye2025dream,
  title={Dream 7b: Diffusion large language models},
  author={Ye, Jiacheng and Xie, Zhihui and Zheng, Lin and Gao, Jiahui and Wu, Zirui and Jiang, Xin and Li, Zhenguo and Kong, Lingpeng},
  journal={arXiv preprint arXiv:2508.15487},
  year={2025},
  url={https://arxiv.org/pdf/2508.15487v1},
}

@misc{nie2025large,
      title={Large Language Diffusion Models}, 
      author={Shen Nie and Fengqi Zhu and Zebin You and Xiaolu Zhang and Jingyang Ou and Jun Hu and Jun Zhou and Yankai Lin and Ji-Rong Wen and Chongxuan Li},
      year={2025},
      eprint={2502.09992},
      archivePrefix={arXiv},
      primaryClass={cs.CL},
      url={https://arxiv.org/abs/2502.09992}, 
}

@misc{zhu2025llada,
      title={LLaDA 1.5: Variance-Reduced Preference Optimization for Large Language Diffusion Models}, 
      author={Fengqi Zhu and Rongzhen Wang and Shen Nie and Xiaolu Zhang and Chunwei Wu and Jun Hu and Jun Zhou and Jianfei Chen and Yankai Lin and Ji-Rong Wen and Chongxuan Li},
      year={2025},
      eprint={2505.19223},
      archivePrefix={arXiv},
      primaryClass={cs.LG},
      url={https://arxiv.org/abs/2505.19223}, 
}

@misc{llada-moe-7b-a1b-base,
      title={LLaDA-MoE: A Sparse MoE Diffusion Language Model}, 
      author={Fengqi Zhu and Zebin You and Yipeng Xing and Zenan Huang and Lin Liu and Yihong Zhuang and Guoshan Lu and Kangyu Wang and Xudong Wang and Lanning Wei and Hongrui Guo and Jiaqi Hu and Wentao Ye and Tieyuan Chen and Chenchen Li and Chengfu Tang and Haibo Feng and Jun Hu and Jun Zhou and Xiaolu Zhang and Zhenzhong Lan and Junbo Zhao and Da Zheng and Chongxuan Li and Jianguo Li and Ji-Rong Wen},
      year={2025},
      eprint={2509.24389},
      archivePrefix={arXiv},
      primaryClass={cs.CL},
      url={https://arxiv.org/abs/2509.24389}, 
}

@misc{cheng2025sdarsynergisticdiffusionautoregressionparadigm,
      title={SDAR: A Synergistic Diffusion-AutoRegression Paradigm for Scalable Sequence Generation}, 
      author={Shuang Cheng and Yihan Bian and Dawei Liu and Linfeng Zhang and Qian Yao and Zhongbo Tian and Wenhai Wang and Qipeng Guo and Kai Chen and Biqing Qi and Bowen Zhou},
      year={2025},
      eprint={2510.06303},
      archivePrefix={arXiv},
      primaryClass={cs.LG},
      url={https://arxiv.org/abs/2510.06303}, 
}

@misc{gong2025diffucoderunderstandingimprovingmasked,
      title={DiffuCoder: Understanding and Improving Masked Diffusion Models for Code Generation}, 
      author={Shansan Gong and Ruixiang Zhang and Huangjie Zheng and Jiatao Gu and Navdeep Jaitly and Lingpeng Kong and Yizhe Zhang},
      year={2025},
      eprint={2506.20639},
      archivePrefix={arXiv},
      primaryClass={cs.CL},
      url={https://arxiv.org/abs/2506.20639}, 
}

@misc{gong2025scalingdiffusionlanguagemodels,
      title={Scaling Diffusion Language Models via Adaptation from Autoregressive Models}, 
      author={Shansan Gong and Shivam Agarwal and Yizhe Zhang and Jiacheng Ye and Lin Zheng and Mukai Li and Chenxin An and Peilin Zhao and Wei Bi and Jiawei Han and Hao Peng and Lingpeng Kong},
      year={2025},
      eprint={2410.17891},
      archivePrefix={arXiv},
      primaryClass={cs.CL},
      url={https://arxiv.org/abs/2410.17891}, 
}

@misc{song2025seeddiffusionlargescalediffusion,
      title={Seed Diffusion: A Large-Scale Diffusion Language Model with High-Speed Inference}, 
      author={Yuxuan Song and Zheng Zhang and Cheng Luo and Pengyang Gao and Fan Xia and Hao Luo and Zheng Li and Yuehang Yang and Hongli Yu and Xingwei Qu and Yuwei Fu and Jing Su and Ge Zhang and Wenhao Huang and Mingxuan Wang and Lin Yan and Xiaoying Jia and Jingjing Liu and Wei-Ying Ma and Ya-Qin Zhang and Yonghui Wu and Hao Zhou},
      year={2025},
      eprint={2508.02193},
      archivePrefix={arXiv},
      primaryClass={cs.CL},
      url={https://arxiv.org/abs/2508.02193}, 
}

@misc{yang2025mmadamultimodallargediffusion,
      title={MMaDA: Multimodal Large Diffusion Language Models}, 
      author={Ling Yang and Ye Tian and Bowen Li and Xinchen Zhang and Ke Shen and Yunhai Tong and Mengdi Wang},
      year={2025},
      eprint={2505.15809},
      archivePrefix={arXiv},
      primaryClass={cs.CV},
      url={https://arxiv.org/abs/2505.15809}, 
}

@misc{you2025lladavlargelanguagediffusion,
      title={LLaDA-V: Large Language Diffusion Models with Visual Instruction Tuning}, 
      author={Zebin You and Shen Nie and Xiaolu Zhang and Jun Hu and Jun Zhou and Zhiwu Lu and Ji-Rong Wen and Chongxuan Li},
      year={2025},
      eprint={2505.16933},
      archivePrefix={arXiv},
      primaryClass={cs.LG},
      url={https://arxiv.org/abs/2505.16933}, 
}

@misc{kim2025anyorderflexiblelengthmasked,
      title={Any-Order Flexible Length Masked Diffusion}, 
      author={Jaeyeon Kim and Lee Cheuk-Kit and Carles Domingo-Enrich and Yilun Du and Sham Kakade and Timothy Ngotiaoco and Sitan Chen and Michael Albergo},
      year={2025},
      eprint={2509.01025},
      archivePrefix={arXiv},
      primaryClass={cs.LG},
      url={https://arxiv.org/abs/2509.01025}, 
}

@misc{bie2025llada2,
      title={LLaDA2.0: Scaling Up Diffusion Language Models to 100B}, 
      author={Tiwei Bie and Maosong Cao and Kun Chen and Lun Du and Mingliang Gong and Zhuochen Gong and Yanmei Gu and Jiaqi Hu and Zenan Huang and Zhenzhong Lan and Chengxi Li and Chongxuan Li and Jianguo Li and Zehuan Li and Huabin Liu and Lin Liu and Guoshan Lu and Xiaocheng Lu and Yuxin Ma and Jianfeng Tan and Lanning Wei and Ji-Rong Wen and Yipeng Xing and Xiaolu Zhang and Junbo Zhao and Da Zheng and Jun Zhou and Junlin Zhou and Zhanchao Zhou and Liwang Zhu and Yihong Zhuang},
      year={2025},
      eprint={2512.15745},
      archivePrefix={arXiv},
      primaryClass={cs.LG},
      url={https://arxiv.org/abs/2512.15745}, 
}

@misc{feng2025theoreticalbenefitlimitationdiffusion,
      title={Theoretical Benefit and Limitation of Diffusion Language Model}, 
      author={Guhao Feng and Yihan Geng and Jian Guan and Wei Wu and Liwei Wang and Di He},
      year={2025},
      eprint={2502.09622},
      archivePrefix={arXiv},
      primaryClass={cs.LG},
      url={https://arxiv.org/abs/2502.09622}, 
}

@misc{liu2025longlladaunlockinglongcontext,
      title={LongLLaDA: Unlocking Long Context Capabilities in Diffusion LLMs}, 
      author={Xiaoran Liu and Zhigeng Liu and Zengfeng Huang and Qipeng Guo and Ziwei He and Xipeng Qiu},
      year={2025},
      eprint={2506.14429},
      archivePrefix={arXiv},
      primaryClass={cs.CL},
      url={https://arxiv.org/abs/2506.14429}, 
}

@misc{wu2025fastdllmtrainingfreeaccelerationdiffusion,
      title={Fast-dLLM: Training-free Acceleration of Diffusion LLM by Enabling KV Cache and Parallel Decoding}, 
      author={Chengyue Wu and Hao Zhang and Shuchen Xue and Zhijian Liu and Shizhe Diao and Ligeng Zhu and Ping Luo and Song Han and Enze Xie},
      year={2025},
      eprint={2505.22618},
      archivePrefix={arXiv},
      primaryClass={cs.CL},
      url={https://arxiv.org/abs/2505.22618}, 
}

@misc{yu2025dimplediscretediffusionmultimodal,
      title={Dimple: Discrete Diffusion Multimodal Large Language Model with Parallel Decoding}, 
      author={Runpeng Yu and Xinyin Ma and Xinchao Wang},
      year={2025},
      eprint={2505.16990},
      archivePrefix={arXiv},
      primaryClass={cs.CV},
      url={https://arxiv.org/abs/2505.16990}, 
}

@misc{liu2025dllmcacheacceleratingdiffusionlarge,
      title={dLLM-Cache: Accelerating Diffusion Large Language Models with Adaptive Caching}, 
      author={Zhiyuan Liu and Yicun Yang and Yaojie Zhang and Junjie Chen and Chang Zou and Qingyuan Wei and Shaobo Wang and Linfeng Zhang},
      year={2025},
      eprint={2506.06295},
      archivePrefix={arXiv},
      primaryClass={cs.LG},
      url={https://arxiv.org/abs/2506.06295}, 
}

@misc{wei2025acceleratingdiffusionlargelanguage,
      title={Accelerating Diffusion Large Language Models with SlowFast Sampling: The Three Golden Principles}, 
      author={Qingyan Wei and Yaojie Zhang and Zhiyuan Liu and Dongrui Liu and Linfeng Zhang},
      year={2025},
      eprint={2506.10848},
      archivePrefix={arXiv},
      primaryClass={cs.CL},
      url={https://arxiv.org/abs/2506.10848}, 
}

@misc{ma2025dinfer,
      title={dInfer: An Efficient Inference Framework for Diffusion Language Models}, 
      author={Yuxin Ma and Lun Du and Lanning Wei and Kun Chen and Qian Xu and Kangyu Wang and Guofeng Feng and Guoshan Lu and Lin Liu and Xiaojing Qi and Xinyuan Zhang and Zhen Tao and Haibo Feng and Ziyun Jiang and Ying Xu and Zenan Huang and Yihong Zhuang and Haokai Xu and Jiaqi Hu and Zhenzhong Lan and Junbo Zhao and Jianguo Li and Da Zheng},
      year={2025},
      eprint={2510.08666},
      archivePrefix={arXiv},
      primaryClass={cs.CL},
      url={https://arxiv.org/abs/2510.08666}, 
}

@misc{peng2025pathplanningmaskeddiffusion,
      title={Path Planning for Masked Diffusion Model Sampling}, 
      author={Fred Zhangzhi Peng and Zachary Bezemek and Sawan Patel and Jarrid Rector-Brooks and Sherwood Yao and Avishek Joey Bose and Alexander Tong and Pranam Chatterjee},
      year={2025},
      eprint={2502.03540},
      archivePrefix={arXiv},
      primaryClass={cs.LG},
      url={https://arxiv.org/abs/2502.03540}, 
}

@misc{kim2025trainworstplanbest,
      title={Train for the Worst, Plan for the Best: Understanding Token Ordering in Masked Diffusions}, 
      author={Jaeyeon Kim and Kulin Shah and Vasilis Kontonis and Sham Kakade and Sitan Chen},
      year={2025},
      eprint={2502.06768},
      archivePrefix={arXiv},
      primaryClass={cs.LG},
      url={https://arxiv.org/abs/2502.06768}, 
}

@misc{huang2025pcsamplerpositionawarecalibrationdecoding,
      title={PC-Sampler: Position-Aware Calibration of Decoding Bias in Masked Diffusion Models}, 
      author={Pengcheng Huang and Shuhao Liu and Zhenghao Liu and Yukun Yan and Shuo Wang and Zulong Chen and Tong Xiao},
      year={2025},
      eprint={2508.13021},
      archivePrefix={arXiv},
      primaryClass={cs.AI},
      url={https://arxiv.org/abs/2508.13021}, 
}

@misc{chen2025dpadefficientdiffusionlanguage,
      title={DPad: Efficient Diffusion Language Models with Suffix Dropout}, 
      author={Xinhua Chen and Sitao Huang and Cong Guo and Chiyue Wei and Yintao He and Jianyi Zhang and Hai "Helen" Li and Yiran Chen},
      year={2025},
      eprint={2508.14148},
      archivePrefix={arXiv},
      primaryClass={cs.CL},
      url={https://arxiv.org/abs/2508.14148}, 
}

@misc{wang2025timefeatureexploitingtemporal,
      title={Time Is a Feature: Exploiting Temporal Dynamics in Diffusion Language Models}, 
      author={Wen Wang and Bozhen Fang and Chenchen Jing and Yongliang Shen and Yangyi Shen and Qiuyu Wang and Hao Ouyang and Hao Chen and Chunhua Shen},
      year={2025},
      eprint={2508.09138},
      archivePrefix={arXiv},
      primaryClass={cs.CL},
      url={https://arxiv.org/abs/2508.09138}, 
}

@misc{cobbe2021trainingverifierssolvemath,
      title={Training Verifiers to Solve Math Word Problems}, 
      author={Karl Cobbe and Vineet Kosaraju and Mohammad Bavarian and Mark Chen and Heewoo Jun and Lukasz Kaiser and Matthias Plappert and Jerry Tworek and Jacob Hilton and Reiichiro Nakano and Christopher Hesse and John Schulman},
      year={2021},
      eprint={2110.14168},
      archivePrefix={arXiv},
      primaryClass={cs.LG},
      url={https://arxiv.org/abs/2110.14168}, 
}

@misc{hendrycks2021measuringmassivemultitasklanguage,
      title={Measuring Massive Multitask Language Understanding}, 
      author={Dan Hendrycks and Collin Burns and Steven Basart and Andy Zou and Mantas Mazeika and Dawn Song and Jacob Steinhardt},
      year={2021},
      eprint={2009.03300},
      archivePrefix={arXiv},
      primaryClass={cs.CY},
      url={https://arxiv.org/abs/2009.03300}, 
}

@inproceedings{dua2019dropreadingcomprehensionbenchmark,
      title={DROP: A Reading Comprehension Benchmark Requiring Discrete Reasoning Over Paragraphs},
      author={Dheeru Dua and Yizhong Wang and Pradeep Dasigi and Gabriel Stanovsky and Sameer Singh and Matt Gardner},
      booktitle={Proceedings of the 2019 Conference of the North American Chapter of the Association for Computational Linguistics: Human Language Technologies},
      year={2019},
      pages={2368--2378},
      publisher={Association for Computational Linguistics},
      doi={10.18653/v1/N19-1246},
      url={https://aclanthology.org/N19-1246},
}

@misc{ma2025korbenchbenchmarkinglanguagemodels,
      title={KOR-Bench: Benchmarking Language Models on Knowledge-Orthogonal Reasoning Tasks}, 
      author={Kaijing Ma and Xinrun Du and Yunran Wang and Haoran Zhang and Zhoufutu Wen and Xingwei Qu and Jian Yang and Jiaheng Liu and Minghao Liu and Xiang Yue and Wenhao Huang and Ge Zhang},
      year={2025},
      eprint={2410.06526},
      archivePrefix={arXiv},
      primaryClass={cs.DB},
      url={https://arxiv.org/abs/2410.06526}, 
}

@misc{chen2021evaluatinglargelanguagemodels,
      title={Evaluating Large Language Models Trained on Code}, 
      author={Mark Chen and Jerry Tworek and Heewoo Jun and Qiming Yuan and Henrique Ponde de Oliveira Pinto and Jared Kaplan and Harri Edwards and Yuri Burda and Nicholas Joseph and Greg Brockman and Alex Ray and Raul Puri and Gretchen Krueger and Michael Petrov and Heidy Khlaaf and Girish Sastry and Pamela Mishkin and Brooke Chan and Scott Gray and Nick Ryder and Mikhail Pavlov and Alethea Power and Lukasz Kaiser and Mohammad Bavarian and Clemens Winter and Philippe Tillet and Felipe Petroski Such and Dave Cummings and Matthias Plappert and Fotios Chantzis and Elizabeth Barnes and Ariel Herbert-Voss and William Hebgen Guss and Alex Nichol and Alex Paino and Nikolas Tezak and Jie Tang and Igor Babuschkin and Suchir Balaji and Shantanu Jain and William Saunders and Christopher Hesse and Andrew N. Carr and Jan Leike and Josh Achiam and Vedant Misra and Evan Morikawa and Alec Radford and Matthew Knight and Miles Brundage and Mira Murati and Katie Mayer and Peter Welinder and Bob McGrew and Dario Amodei and Sam McCandlish and Ilya Sutskever and Wojciech Zaremba},
      year={2021},
      eprint={2107.03374},
      archivePrefix={arXiv},
      primaryClass={cs.LG},
      url={https://arxiv.org/abs/2107.03374}, 
}

@misc{jain2024livecodebenchholisticcontaminationfree,
      title={LiveCodeBench: Holistic and Contamination Free Evaluation of Large Language Models for Code}, 
      author={Naman Jain and King Han and Alex Gu and Wen-Ding Li and Fanjia Yan and Tianjun Zhang and Sida Wang and Armando Solar-Lezama and Koushik Sen and Ion Stoica},
      year={2024},
      eprint={2403.07974},
      archivePrefix={arXiv},
      primaryClass={cs.SE},
      url={https://arxiv.org/abs/2403.07974}, 
}

@misc{hendrycks2021measuringmathematicalproblemsolving,
      title={Measuring Mathematical Problem Solving With the MATH Dataset}, 
      author={Dan Hendrycks and Collin Burns and Saurav Kadavath and Akul Arora and Steven Basart and Eric Tang and Dawn Song and Jacob Steinhardt},
      year={2021},
      eprint={2103.03874},
      archivePrefix={arXiv},
      primaryClass={cs.LG},
      url={https://arxiv.org/abs/2103.03874}, 
}

@misc{pytorch20,
  title   = {PyTorch 2.0: Our Next Generation 2.0 Release},
  author  = {Jain, Animesh and Zhang, Shunting and Yang, Edward and others},
  journal = {arXiv preprint arXiv:2305.01916},
  year    = {2023},
  url     = {https://arxiv.org/abs/2305.01916}
}

@inproceedings{kwon2023vllm,
      title={Efficient Memory Management for Large Language Model Serving with PagedAttention},
      author={Woosuk Kwon and Zhuohan Li and Siyuan Zhuang and Ying Sheng and Lianmin Zheng and Cody Hao Yu and Joseph E. Gonzalez and Hao Zhang and Ion Stoica},
      booktitle={Proceedings of the 29th Symposium on Operating Systems Principles},
      year={2023},
      pages={611--626},
      publisher={Association for Computing Machinery},
      doi={10.1145/3600006.3613165},
      url={https://dl.acm.org/doi/10.1145/3600006.3613165},
}
\appendix
\section{Statements}
In this section, we provide clarifications on issues that require explicit statements. The Ethics Statement is omitted, as this work does not raise any ethical concerns.

\subsection{Reproducibility.} We employ four widely adopted and advanced dLLM models, LLaDA-8B-Instruct, LLaDA-MoE-Instruct, LLaDA2-Mini and LLaDA2-Flash, along with their publicly available weights. Detailed experimental configurations for each setup are provided, and the specific metric configurations for score evaluation are included in Appendix \ref{AppendixB:Eval Cfg}. We utilize TPF, a metric that is relatively robust to confounding factors, as it is not influenced by the number of GPUs, hardware model, or inference framework. Therefore, we argue that the results reported in this paper are highly reproducible.
\subsection{Use of LLMs}
In this study, LLMs were used only to check and improve the accuracy, coherence, and academic quality of the writing. We confirm that LLMs were not employed to generate any experimental data or substantive content. All data originate from our own experiments, and all analyses and conclusions represent the authors’ original work.

\section{Evaluation Config}
\begin{table}[h]
\footnotesize
\raggedright
\caption{Benchmarks, their corresponding evaluation metrics, and In-Context Learning (ICL) setup.}
\renewcommand{\arraystretch}{1.2}
\begin{tabular}{@{} l l c}
\toprule
\textbf{Benchmark} &\textbf{Metric} &\textbf{ICL} \\
\midrule
GSM8K                          &Accuracy                &0-shot \\
Math                           &Accuracy                &0-shot \\
SQuAD2.0                       &Score                   &1-shot \\
DROP                           &Score                   &2-shot \\
MMLU                  &Weighted Average       &0-shot \\
KorBench                       &Naive Average          &0-shot \\
LCB OC Code Generation v6 &Score                   &0-shot \\
OpenAI HumanEval               &Pass@1  &0-shot \\
\bottomrule
\end{tabular}
\label{tab:benchmark_metric}
\end{table}
\label{AppendixB:Eval Cfg}

We employ OpenCompass to assist in the evaluation process, ensuring a standardized and systematic assessment. For each benchmark (LCB for LiveCodeBench), we use the specific metrics presented in Table~\ref{tab:benchmark_metric}, which allow for a consistent and comprehensive comparison across different tasks. Regarding in-context learning (ICL) configurations, all benchmarks were evaluated in a zero-shot setting, except for Drop and SQuAD2, which were evaluated in two-shot and one-shot settings respectively.

\section{CreditDecoding Supplement}
In this section, we provide additional details and experiments related to CreditDecoding that were not presented in the main body of the paper.

\subsection{Deriving the Minimum Logit Gain }
\label{AppendixC1:appendix_xmin}

In this subsection, we derive the minimum gain $X$ induced by adding $\log X$ to a token logit, such that the boosted token probability exceeds the decoding threshold.

Consider a fixed position $i$ at denoising step $t$ with logits $\{l_t^{i,u}\}_{u\in\mathcal V}$ and the corresponding softmax distribution
\begin{equation}
p_t^{i,u}=\frac{e^{l_t^{i,u}}}{\sum_{z\in\mathcal V} e^{l_t^{i,z}}}.
\label{eq:softmax_def}
\end{equation}
Let $v\in\mathcal V$ be the token to be boosted. We apply a logit gain:
\begin{equation}
\hat{l}_t^{i,u}=
\begin{cases}
l_t^{i,u}+\log X, & u=v,\\
l_t^{i,u}, & u\neq v,
\end{cases}
\qquad X\ge 1,
\label{eq:logit_gain_appendix}
\end{equation}
and define the enhanced distribution by softmax:
\begin{equation}
\hat{p}_t^{i,u}=\frac{e^{\hat{l}_t^{i,u}}}{\sum_{z\in\mathcal V} e^{\hat{l}_t^{i,z}}}.
\label{eq:softmax_hat_def}
\end{equation}
Substituting Eq.~\ref{eq:logit_gain_appendix} into Eq.~\ref{eq:softmax_hat_def}, we obtain:
\begin{equation}
\label{eq:phat_logits}
\begin{split}
\hat{p}_t^{i,v}
&= \frac{e^{l_t^{i,v}+\log X}}{e^{l_t^{i,v}+\log X}+\sum_{u\neq v}e^{l_t^{i,u}}} \\
&= \frac{X e^{l_t^{i,v}}}{X e^{l_t^{i,v}}+\sum_{u\neq v}e^{l_t^{i,u}}} \\
&= \frac{X p_t^{i,v}}{(1-p_t^{i,v}) + X p_t^{i,v}}.
\end{split}
\end{equation}
To decode token $v$ under threshold $\tau\in(0,1)$, we require $\hat{p}_t^{i,v}\ge \tau$. By Eq.~\ref{eq:phat_logits}:
\begin{equation}
\label{eq:xmin_appendix}
X \ge X_{\min}(p_t^{i,v},\tau) = \frac{\tau}{1-\tau}\cdot\frac{1-p_t^{i,v}}{p_t^{i,v}}.
\end{equation}

Consequently, for under-confident tokens (e.g., $p_t^{i,v}\approx 1/|\mathcal V|$), the required credit $X$ becomes prohibitively large, validating that naive gain application is risky in early denoising steps with unstable probabilities.

\subsection{Algorithm}
\begin{algorithm}[t]
\footnotesize
\caption{CreditDecoding}
\label{alg:creditdecode}
\begin{algorithmic}[1]
\State \textbf{Input:} denoiser $f_\theta$, threshold $\tau$, fusion strength $\alpha$, decay $\beta$, boost exponent $\gamma$
\State Initialize $x_T \gets (\mask,\dots,\mask)$

\For{each decoding block $[b_s,b_e)$}
    \State Initialize credit tensor $C \gets \mathbf{0}$ \Comment{Per-block credit}
    \For{$t = T, T-1, \dots, 1$} \Comment{Denoising}
        \State $x_t^{\text{blk}} \gets x_t[b_s{:}b_e]$
        \State $m \gets \mathbb{I}\big[x_t^{\text{blk}} = \mask\big]$

        \Comment{\textit{\textbf{Prediction and Greedy selection}}}
        \State $\ell \gets f_\theta(x_t)_{b_s{:}b_e}$
        \State $p \gets \mathrm{Softmax}(\ell)$
        \State $(p^{\max}, v^{\max}) \gets \max_{v\in\mathcal V} p(\cdot)$
        \State $\Delta \gets (p^{\max})^\gamma \odot m$

        \Comment{\textit{\textbf{Credit Update}}}
        \State $C \gets \beta \cdot C$ \Comment{Global decay}
        \State $C \gets C + \mathrm{ScatterAdd}(C,\, v^{\max},\, \Delta)$ \Comment{Focused enhancement}
        \State $\hat{\ell} \gets \ell + \alpha \cdot \log(1 + C)$ \Comment{Logits fusion}

        \Comment{\textit{\textbf{Threshold-based Decoding}}}
        \State $\hat{p} \gets \mathrm{Softmax}(\hat{\ell})$
        \State $(s^{\max},\, \tilde{x}) \gets \max_{v\in\mathcal V} \hat{p}(\cdot)$\Comment{Predicted tokens and  confidence}
        \State $\mathrm{idx} \gets (s^{\max} \ge \tau) \land m$ \Comment{Positions to decode}

        \Comment{\textit{\textbf{Update block state}}}
        \State $x_{t-1}[b_s{:}b_e] \gets \mathrm{where}(\mathrm{idx},\, \tilde{x},\, x_t^{\text{blk}})$
    \EndFor
\EndFor
\State \textbf{return} $x_0$
\end{algorithmic}
\end{algorithm}

To accurately illustrate the workflow of the CreditDecoding algorithm, we present its detailed steps in Algorithm~\ref{alg:creditdecode}. Here, 
$T$ denotes the total number of decoding steps, and $r$,$s$,$t$ correspond to the mask ratios of the previous, current, and subsequent steps, respectively. Lines 4–12 describe the process of trace credit accumulation, where historical trace credit are decayed by a factor $\beta$ and the token with the highest confidence is assigned additional trace credit. Lines 13–14 show the application of trace credit, in which past information is integrated into the current step by fusing it with the original logits. Finally, Lines 16–23 represent the mainstream approach to parallel decoding, namely the thresholding method, whose effectiveness has been demonstrated in Fast-dLLM~\citep{wu2025fastdllmtrainingfreeaccelerationdiffusion}.

\subsection{Ideal Decoding Boundary}
\label{AppendixC3:Ideal Decoding Boundary}
\begin{figure}[h]
    \centering
    \includegraphics[width=0.9\linewidth]{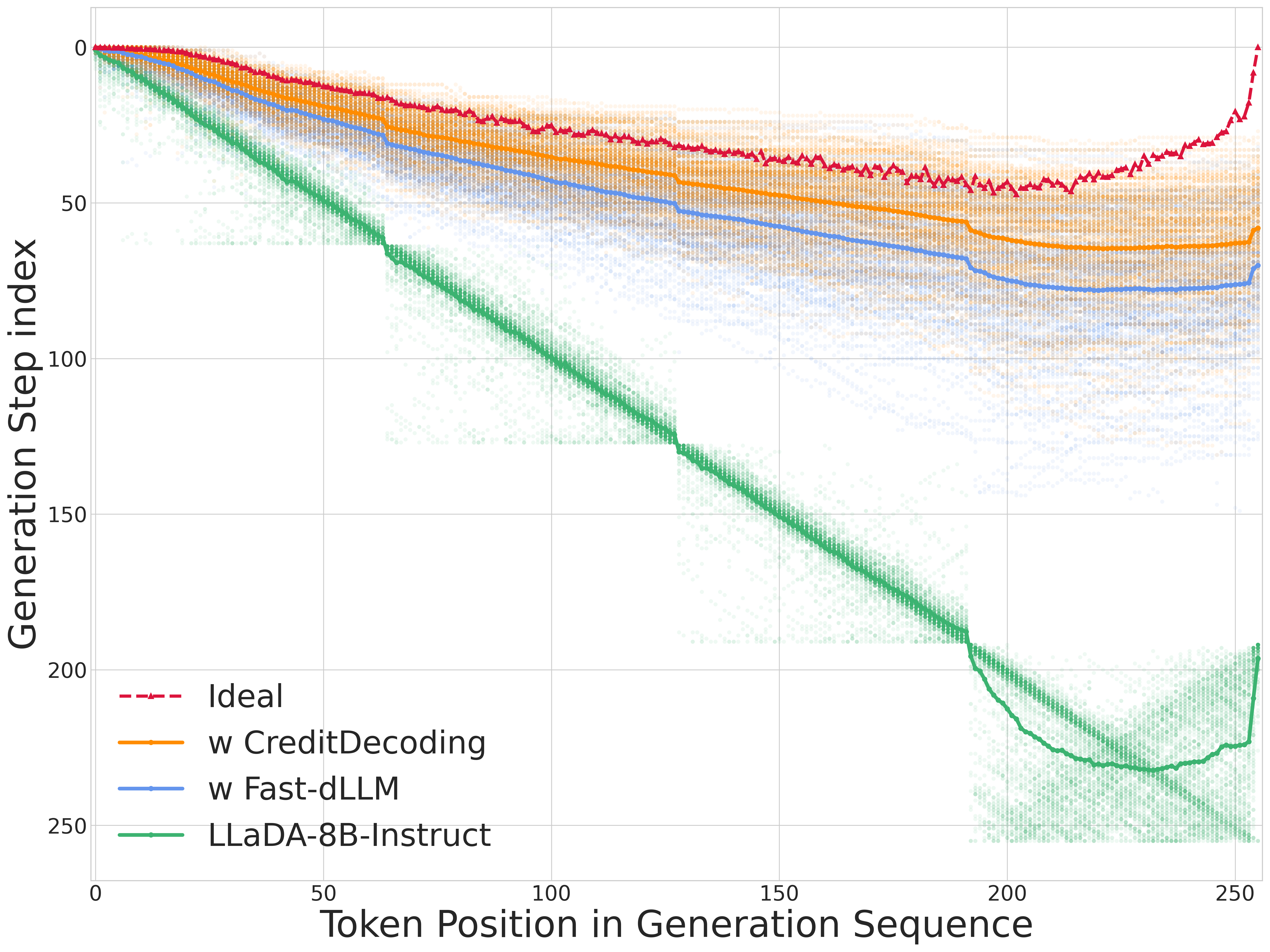}
    \caption{
    \textbf{Decoding Boundary between token stabilization and final decoding.} Evaluated on GSM8K using LLaDA-8B-Instruct, the \textit{red} curve represents the theoretical decoding bound, while others show actual decoding bound. The persistent gap highlights the computational redundancy where correct tokens are repeatedly remasked despite early stabilization.}
    \label{fig:temperal_gap}
\vspace{-10pt}
\end{figure}

In Figure~\ref{fig:temperal_gap}, we illustrate the decoding boundaries of three methods, LLaDA, Fast-dLLM, and CreditDecoding, within a single plot. Figure~\ref{fig5:rank_correct} further presents the decoding boundaries (highlighted by red lines) from the perspective of confidence rank, along with an analysis of the ideal decoding boundary. For a more intuitive comparison, the ranges of the plots in Figure~\ref{fig:temperal_gap} are aligned, whereas the vertical scales of the three subplots in Figure~\ref{fig5:rank_correct} differ to emphasize the detailed variations introduced by thresholding. 
The line connecting the first appearance of yellow points marks where target tokens are initially predicted.

It is worth noting that, as shown in Figure~\ref{fig5:rank_correct}, various methods yield distinct ideal decoding boundaries. This variation arises because each method receives distinct inputs at every step, resulting in different confidence distributions. Importantly, stronger models or more effective methods tend to produce an ideal decoding boundary that shifts upward. Our CreditDecoding approach, however, is orthogonal to most existing methods and can further improve their ideal decoding boundaries, bringing the actual decoding boundary closer to the ideal one.

\begin{figure*}[h]
    \centering
    \begin{minipage}{0.45\linewidth}
        \includegraphics[width=\linewidth]{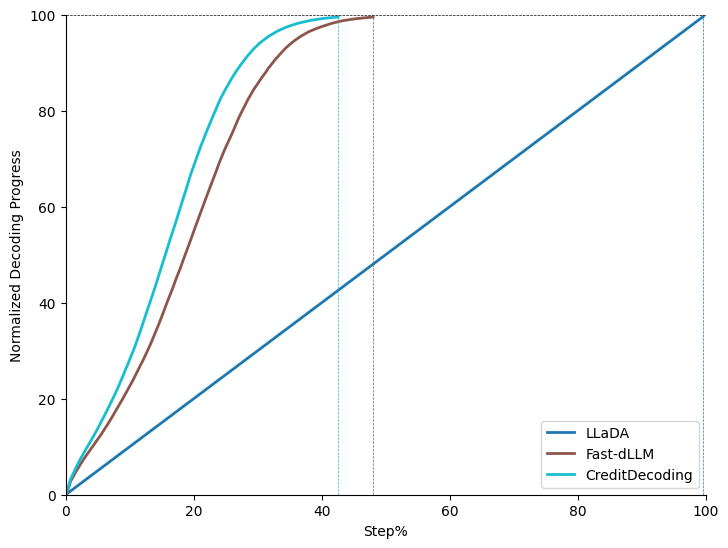}
        \subcaption{Normalized Decoding Progress on GSM8K}\label{fig:tpf_subfig1}
    \end{minipage} \hfill
    \begin{minipage}{0.45\linewidth}
        \includegraphics[width=\linewidth]{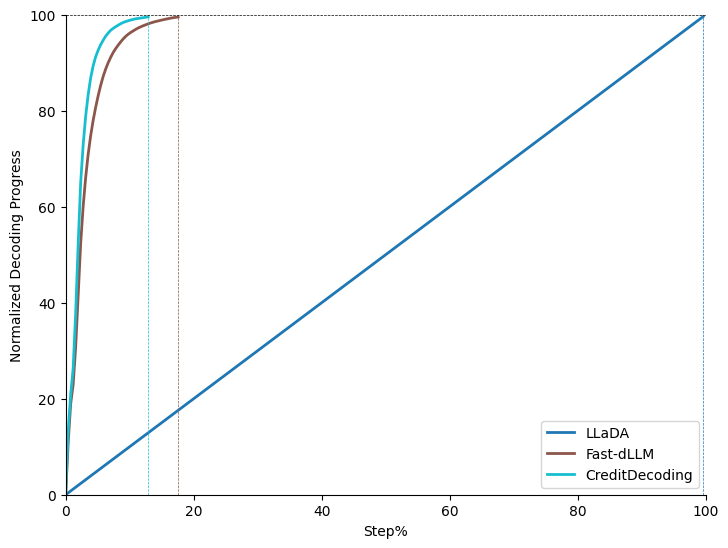}
        \subcaption{Normalized Decoding Progress on SQuAD2.0}\label{fig:tpf_subfig2}
    \end{minipage} \hfill
    \caption{
    Normalized Decoding Progress \textbf{w/o Early Stop} on \textbf{GSM8K} and \textbf{SQuAD2.0}. We demonstrate the \emph{decoding progress} through visualizing the accumulated number of decoded tokens per step, using \textbf{LLaDA-8B-Instruct} with \textit{generation length $=256$} and \textit{block size $=64$}. In order to ensure comparability of the decoding progress, we did not use early stopping. The vertical dashed lines in the figure mark the denoising step at which each method completes, and the ratio of these steps represents the speedup.}
    \label{fig:normalized_tpf}
\end{figure*}

In Figure~\ref{fig:normalized_tpf}, we visualize the accumulated decoded tokens per step for LLaDA, Fast-dLLM, and CreditDecoding on specific datasets. Two key observations emerge: First, the speedup of the same method varies across datasets. Second, CreditDecoding consistently outperforms Fast-dLLM, especially on SQuAD2.0, where it reduces the denoising steps by an additional 5\% compared to Fast-dLLM's 20\% reduction. As an orthogonal method, CreditDecoding offers greater improvements with higher speedups.

\subsection{Block Length Ablation}
\label{AppendixC4:block_length}
\begin{figure}[h]
    \centering
    \includegraphics[width=\linewidth]{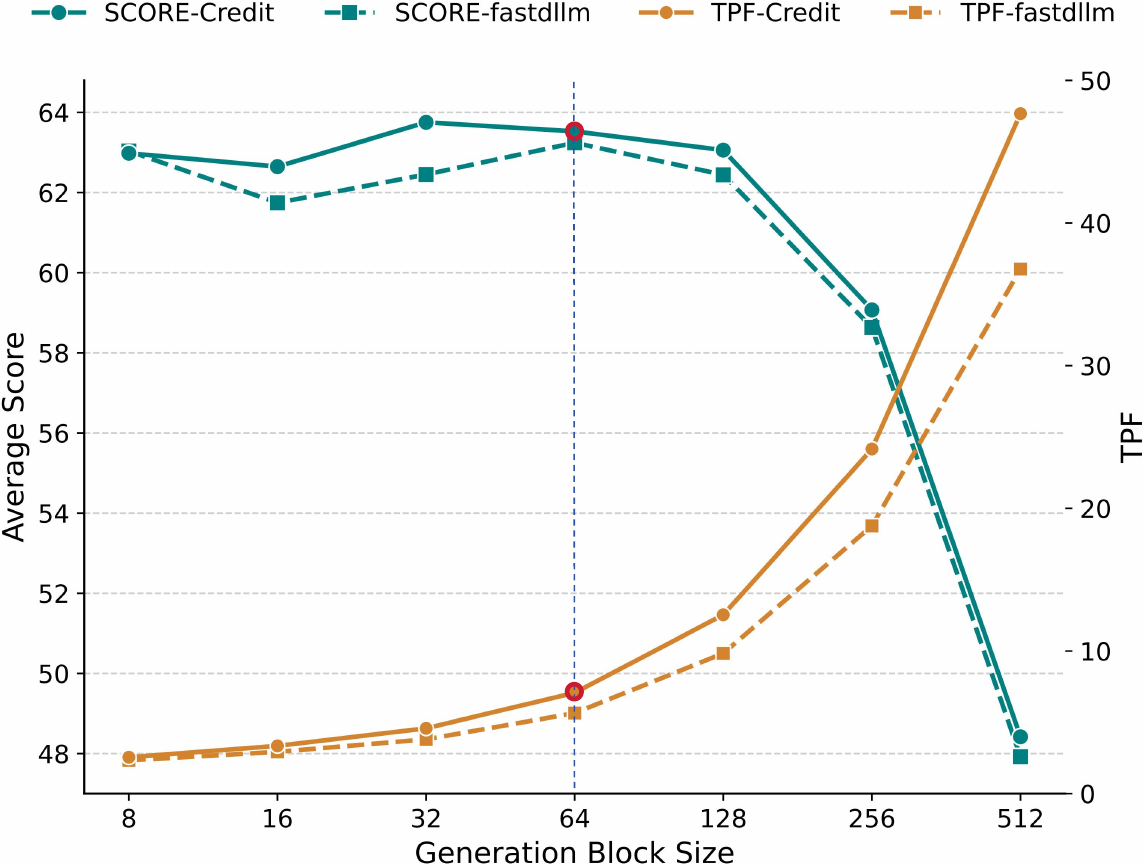}
    \caption{Block Length Ablation Study}
    \label{fig:block_ablation}
\end{figure}
In the parallel decoding framework of generative models, the choice of generation block length directly determines the balance between system performance and efficiency. Through block length ablation experiments analyzing Fast-dLLM and CreditDecoding, as shown in Figure~\ref{fig:block_ablation}, it can be observed that when the block length is set to 64, both methods demonstrate optimal balance. Under this configuration, both CreditDecoding and Fast-dLLM achieve higher average scores, indicating that this scale fully unleashes the potential of block-level generation mechanisms without significantly sacrificing generation quality. In contrast, smaller block lengths (such as 32) can maintain high performance levels but result in significantly lower inference speeds, making it difficult to meet the high-throughput requirements of practical deployment. Meanwhile, larger block lengths (such as 256 and above) significantly improve TPF but lead to a sharp performance decline. Therefore, a block length of \textbf{64} is an ideal choice considering both accuracy and practicality.

We also observe that the performance degradation with increasing block size is caused by the parallel decoding mechanism itself, as dLLMs inevitably suffer from prematurely decoded tokens. This issue also appears in other training-free methods like Top-$K$ or entropy-based thresholding. Nevertheless, CreditDecoding significantly alleviates this degradation. By capturing \textit{the temporal consistency of confidence} during denoising, CreditDecoding effectively filters out noise from premature predictions, thereby better maintaining performance while accelerating compared to Fast-dLLM.

A closer comparison in Figure~\ref{fig:block_ablation} reveals that CreditDecoding demonstrates greater advantages in overall performance. In terms of performance, CreditDecoding reaches its peak at a block length of 32 and consistently performs on par with or better than Fast-dLLM across medium and small scales. Even under large block lengths where both methods decline, CreditDecoding still exhibits better error control. Regarding speed, CreditDecoding shows higher TPF across all scales, and the performance gap widens as block length increases, reflecting its architectural advantages in parallel scenarios.

In summary, CreditDecoding's core value lies in its superior generation performance compared to Fast-dLLM, along with higher overall efficiency. Future work could explore adaptive block sizing or hybrid decoding strategies to dynamically adjust block length and further optimize the balance between performance and speed across different sequence lengths.

\begin{figure}[t]
    \centering
    \includegraphics[width=\linewidth]{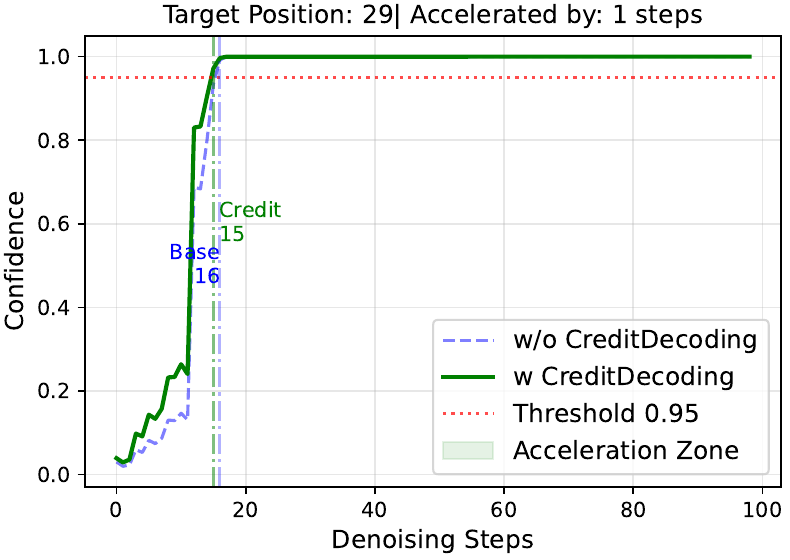}
    \caption{A case study comparing the confidence evolution of a target token under Fast-dLLM and CreditDecoding. It illustrates the direct impact of trace credit on a single token.
    While this local gain appears minor, it iteratively improves the context for subsequent tokens, which ultimately yields a $21.3\%$ reduction in total inference steps.
    }
    \label{fig:toy_study}
    \vspace{-10pt}
\end{figure}

\subsection{Contextual Cumulative Effect}
\label{AppendixC5:Contextual Cumulative Effect}
\begin{figure*}
    \centering
    \begin{minipage}{0.45\linewidth}
        \centering
        \includegraphics[width=\linewidth]{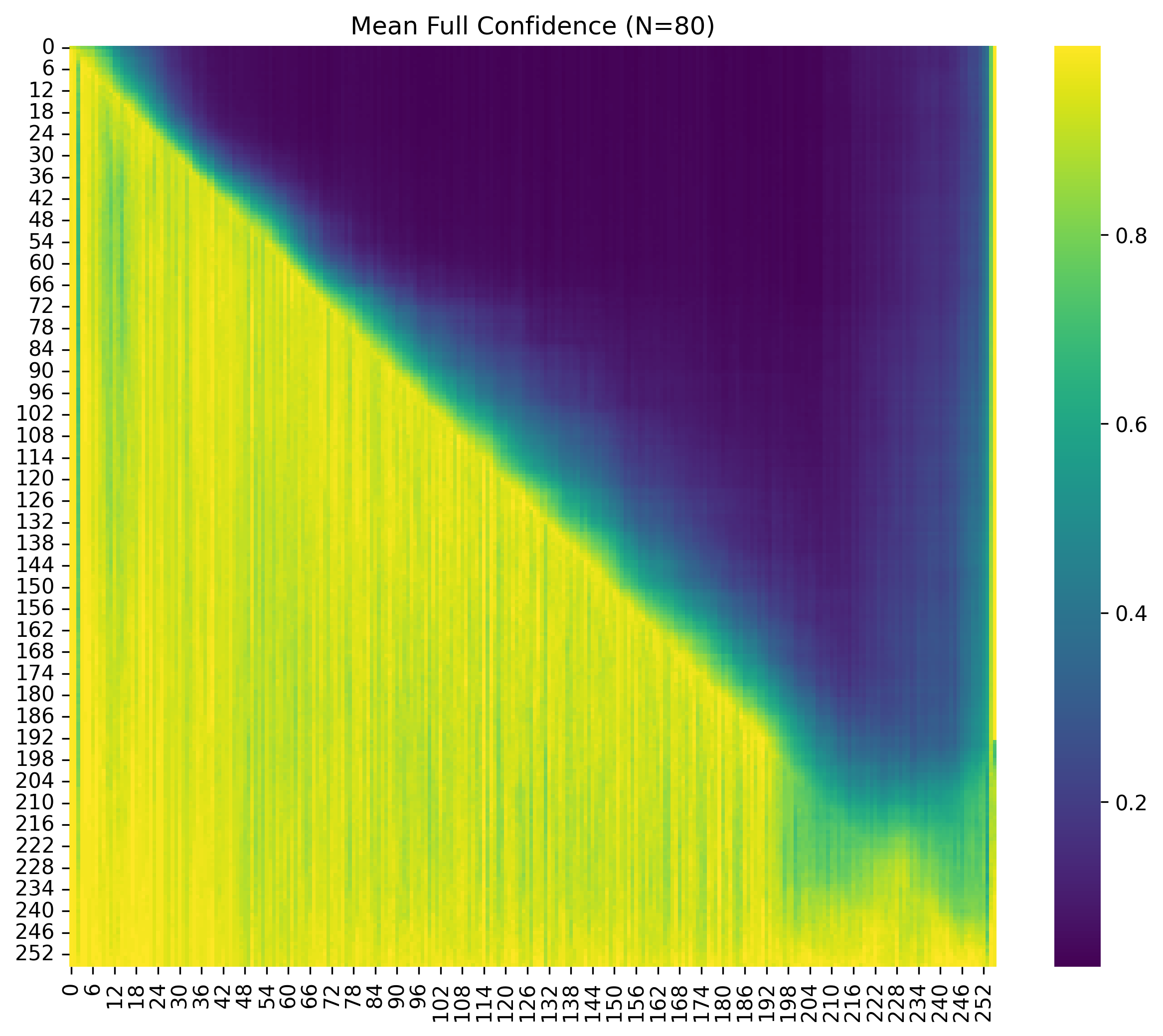}
        \subcaption{Token Confidence(GSM8K)}\label{fig:subfig1}
        \label{fig:gsm_confdc}
    \end{minipage} \hfill
    \begin{minipage}{0.45\linewidth}
        \centering
        \includegraphics[width=\linewidth]{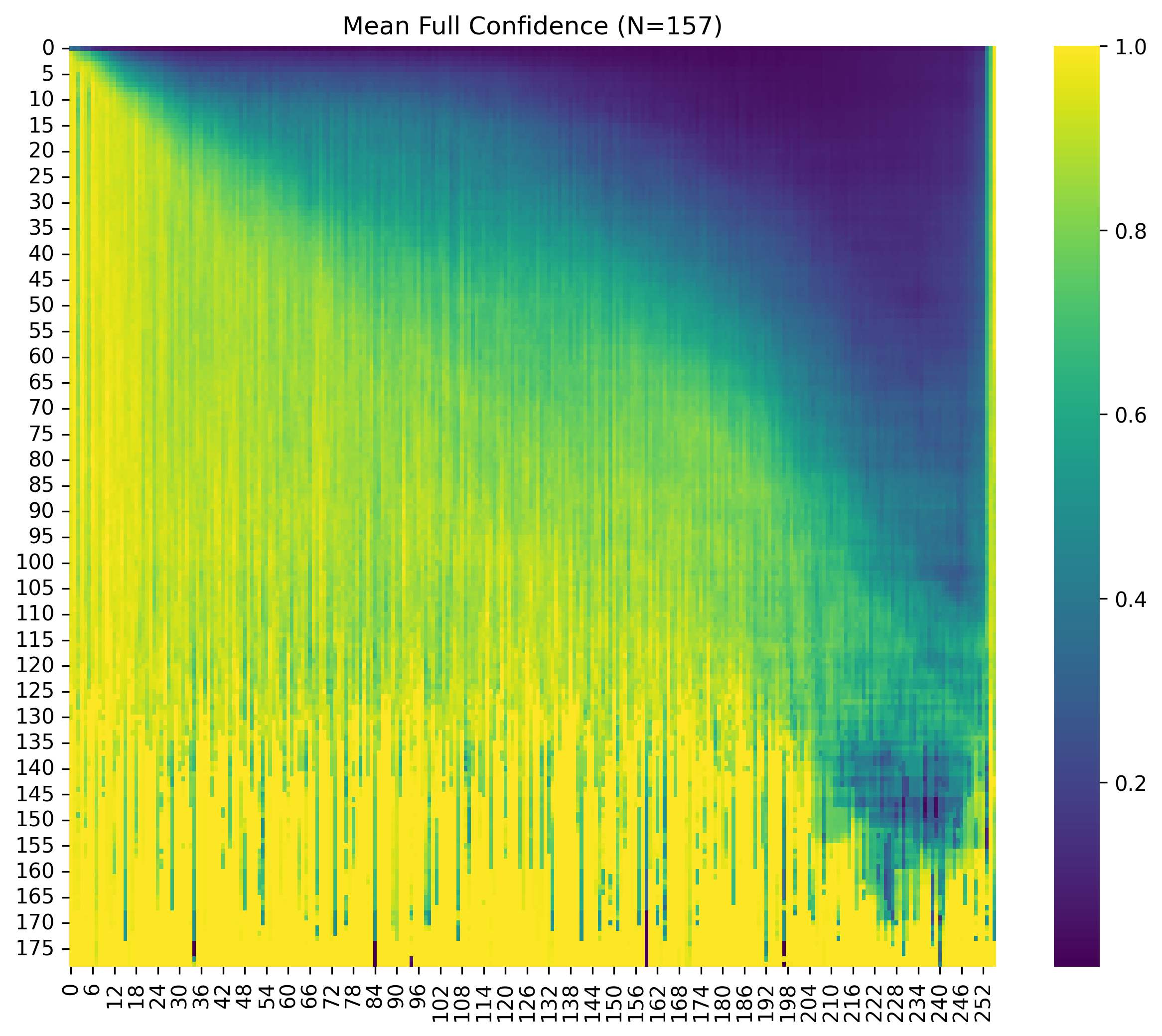}
        \subcaption{Token Confidence(HumanEval)}\label{fig:subfig2}
    \end{minipage}
    
    \vspace{0.2cm} 
    
    \begin{minipage}{\linewidth}
        \centering
        \includegraphics[width=\linewidth]{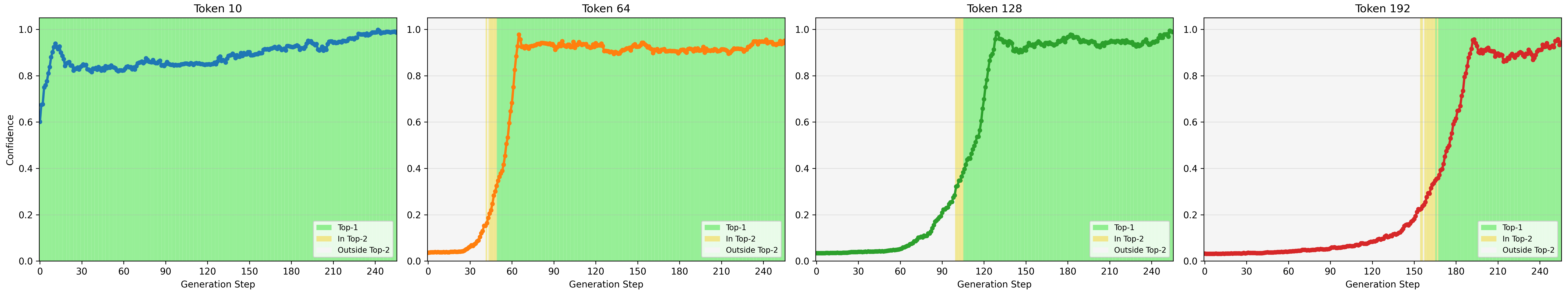}
        \subcaption{Baseline Confidence Convergence(GSM8K)}\label{fig:subfig3}
    \end{minipage} \hfill
    \begin{minipage}{\linewidth}
        \centering
        \includegraphics[width=\linewidth]{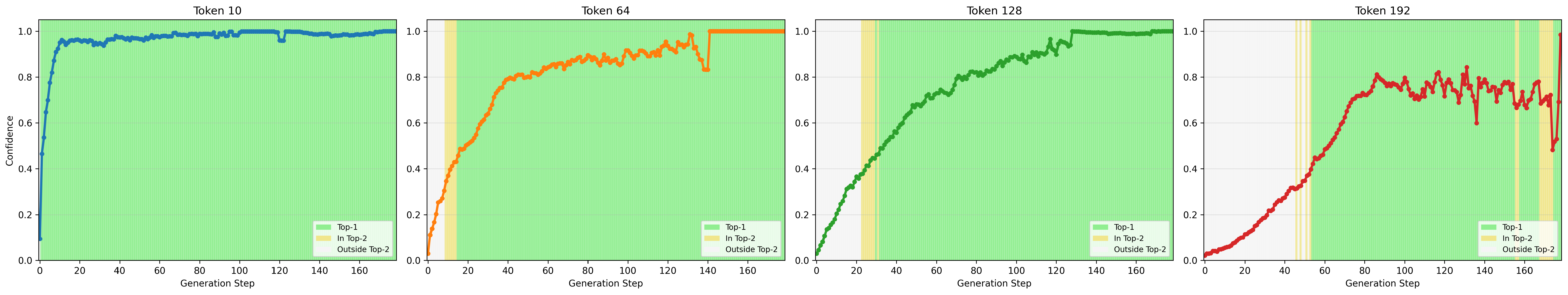}
        \subcaption{Baseline Confidence Conv.(HumanEval)}\label{fig:subfig4}
    \end{minipage}
    
    \vspace{0.2cm} 
    
    \begin{minipage}{\linewidth}
        \centering
        \includegraphics[width=\linewidth]{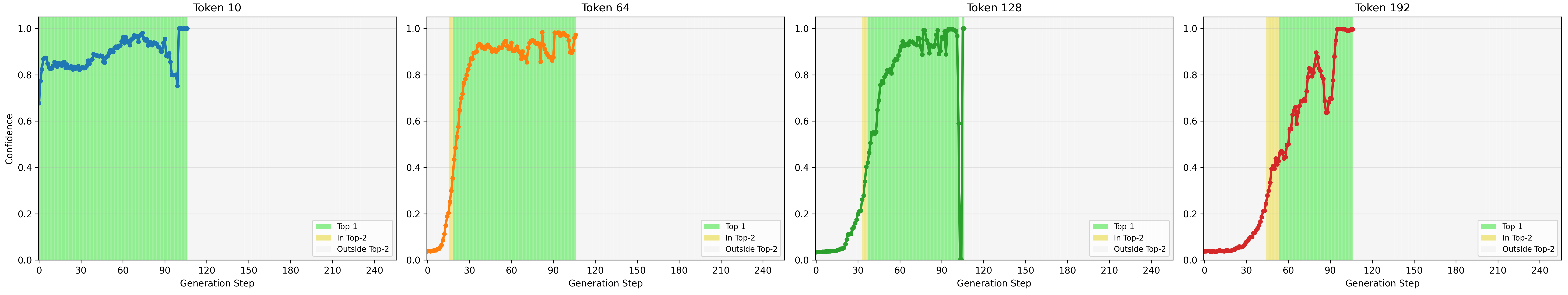}
        \subcaption{Credit Confidence Conv(GSM8K)}\label{fig:subfig5}
    \end{minipage} \hfill
    \begin{minipage}{\linewidth}
        \centering
        \includegraphics[width=\linewidth]{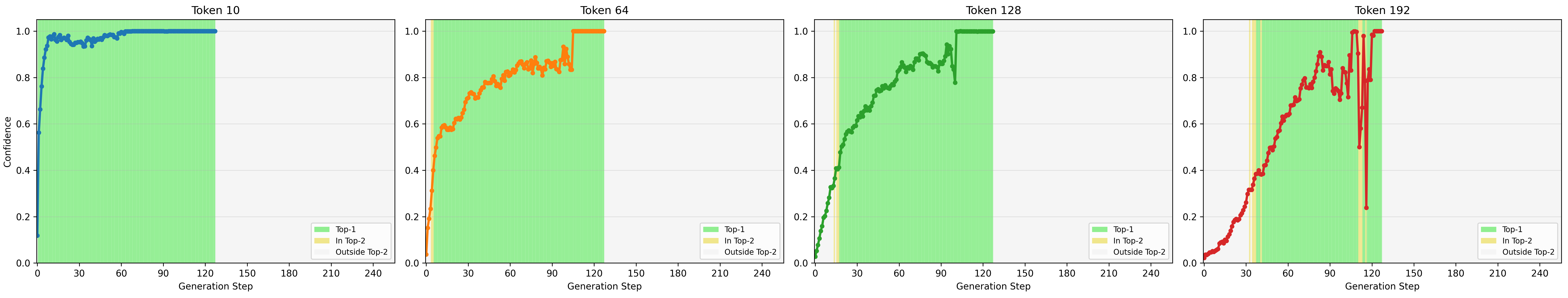}
        \subcaption{Credit Confidence Conv.(HumanEval)}\label{fig:subfig6}
    \end{minipage}
    
    \caption{Token confidence across datasets. Figures (a) and (b) present the average confidence of $N$ tokens at each position during inference. Figures (c)--(f) illustrate the convergence behavior of four representative tokens located at positions 10, 64, 128, and 192, respectively. Results in (a), (c), and (e) are sampled from GSM8K, while those in (b), (d), and (f) are from HumanEval. Panels (c) and (d) show baseline convergence, and (e) and (f) show convergence with CreditDecode.}
    \label{fig:token confidence}
\end{figure*}

Notably, CreditDecoding does not merely accelerate the decoding of an individual token; by decoding some tokens earlier, it improves the context for subsequent denoising steps and can increase the confidence of other still-masked positions as well.
Figure~\ref{fig:toy_study} provides a representative example: although CreditDecoding reduces the number of denoising steps, the confidence trajectory of the target token remains largely unchanged, suggesting that the speedup mainly stems from earlier decoding enabled by improved intermediate context, rather than altering the model’s intrinsic predictions.

\subsection{Dataset-Dependent Denoising Traces}
\label{AppendixC6:dataset-dependent}

Further analysis in Figure~\ref{fig:token confidence} compares GSM8K and HumanEval. GSM8K requires more context from prior steps, leading to delayed but sharp confidence gains, while HumanEval shows earlier, gradual confidence increases. With CreditDecoding, HumanEval retains its confidence trend while reducing inference steps, whereas GSM8K experiences minor fluctuations that contribute to the performance drop reported in Table~\ref{tab:Main-Results}.

In Figures~\ref{fig:token confidence} (c) and (d), we observe the confidence convergence of four specific tokens. On GSM8K, the confidence increase is delayed for tokens later in the sequence, indicating that later tokens rely on earlier tokens to build confidence. Once earlier tokens are decoded, confidence increases rapidly. On HumanEval, confidence increases earlier in the sequence but at a slower rate.

In Figures~\ref{fig:token confidence} (e) and (f), after applying CreditDecoding, we observe that for HumanEval, CreditDecoding retains the original confidence trends while reducing the total inference steps, leading to faster inference without loss of accuracy. For GSM8K, early predictions introduce slight confidence fluctuations, contributing to the performance degradation observed in Table~\ref{tab:Main-Results}.

CreditDecoding essentially reduces the inference complexity by leveraging past judgments, improving both accuracy and efficiency. Its goal is to approach the yellow-blue boundary in Figures (a) and (b), representing the theoretical limit of acceleration. As more powerful base models emerge, the acceleration ceiling of CreditDecoding will increase, leading to further efficiency gains in model inference.

\subsection{Orthogonality}
\label{AppendixC7:Orthogonality}
\newcommand{\cgreen}[1]{\ensuremath{_{\textcolor{green!50!black}{#1}}}}
\newcommand{\cred}[1]{\ensuremath{_{\textcolor{red!50!black}{#1}}}}

\begin{table}[t]
\centering
\caption{Orthogonality analysis of CreditDecoding (CD) combined with various optimizations. The baseline setting is \textbf{w/o Early Stop}. Values in parenthesis/subscript denote the improvement brought by CD.}
\label{tab:Orthogonality_Combined}
\renewcommand{\arraystretch}{0.95}
\setlength{\tabcolsep}{2.5pt}

\resizebox{\columnwidth}{!}{
\begin{tabular}{lccc}
\toprule
\textbf{Method} & \textbf{TPS} & \textbf{TPF} & \textbf{Score} \\
\midrule
\multicolumn{4}{c}{\textit{\textbf{LLaDA-8B-Instruct}}} \\
\midrule
Standard Inference & 7.95 & 1.00 & 60.12 \\
\quad \textit{+ CreditDecoding} & 34.90\cgreen{+339\%} & 15.39\cgreen{+1439\%} & 60.73\cgreen{+0.61} \\
\addlinespace[3pt]
Fast-dLLM (w/o KV) & 28.90 & 12.64 & 60.11 \\
\quad \textit{+ CreditDecoding} & 34.90\cgreen{+21\%} & 15.39\cgreen{+22\%} & 60.73\cgreen{+0.62} \\
\addlinespace[3pt]
Fast-dLLM (w/ KV) & 39.38 & 4.42 & 58.51 \\
\quad \textit{+ CreditDecoding} & 51.40\cgreen{+31\%} & 14.00\cgreen{+217\%} & 58.63\cgreen{+0.12} \\
\addlinespace[3pt]
Early Stop & 9.70 & 1.00 & 59.94 \\
\quad \textit{+ CreditDecoding} & 37.33\cgreen{+285\%} & 6.98\cgreen{+598\%} & 60.75\cgreen{+0.81} \\
\addlinespace[3pt]
PyTorch Compiler & 9.03 & 1.00 & 60.26 \\
\quad \textit{+ CreditDecoding} & 39.51\cgreen{+337\%} & 15.41\cgreen{+1441\%} & 60.43\cgreen{+0.17} \\

\midrule
\multicolumn{4}{c}{\textit{\textbf{LLaDA-MoE-Instruct}}} \\
\midrule
Standard Inference & 3.53 & 1.00 & 62.73 \\
\quad \textit{+ CreditDecoding} & 15.26\cgreen{+333\%} & 14.80\cgreen{+1380\%} & 62.97\cgreen{+0.24} \\
\addlinespace[3pt]
Fast-dLLM (w/o KV) & 13.30 & 13.08 & 62.74 \\
\quad \textit{+ CreditDecoding} & 15.26\cgreen{+15\%} & 14.80\cgreen{+13\%} & 62.97\cgreen{+0.23} \\
\addlinespace[3pt]
FP8 Quantization & 2.72 & 1.00 & 62.47 \\
\quad \textit{+ CreditDecoding} & 11.87\cgreen{+336\%} & 14.45\cgreen{+1345\%} & 62.58\cgreen{+0.11} \\
\addlinespace[3pt]
Early Stop & 5.11 & 1.00 & 62.66 \\
\quad \textit{+ CreditDecoding} & 16.99\cgreen{+233\%} & 4.77\cgreen{+377\%} & 62.92\cgreen{+0.26} \\
\addlinespace[3pt]
PyTorch Compiler & 2.42 & 1.00 & 63.29 \\
\quad \textit{+ CreditDecoding} & 10.72\cgreen{+343\%} & 14.88\cgreen{+1388\%} & 62.99\cred{-0.3} \\
\bottomrule
\end{tabular}
}
\end{table}
\begin{table}[t]
\centering
\caption{
Main benchmark results \textbf{w/ Early Stop} across eight datasets on \textbf{LLaDA-8B-Instruct}
(Gen Length=256, Block Size=64). Cells show \textbf{Score} (top, relative to LLaDA) and \textbf{TPF} (bottom, improvement over Fast-dLLM).
}
\label{tab:Main-Results-8B}
\renewcommand{\arraystretch}{0.95}
\setlength{\tabcolsep}{2.5pt}

\resizebox{0.95\columnwidth}{!}{
\begin{tabular}{lccc}
\toprule
\textbf{Benchmark} ${}^{^{\textbf{Score}}}_{_{\textbf{TPF}}}$ & \textbf{Fast-dLLM} & \textbf{CD} & \textbf{CD}$^{\dagger}$ \\
\midrule
MMLU
& 62.43\cred{-0.03} & \textbf{63.78}\cgreen{+1.32} & 63.66\cgreen{+1.20} \\
& 2.86 & \textbf{4.57}\cgreen{+56\%} & 3.38\cgreen{+18\%} \\
\addlinespace[3pt]

SQuAD2.0
& 91.43\cgreen{+0.00} & \textbf{91.71}\cgreen{+0.28} & 91.48\cgreen{+0.05} \\
& 13.55 & \textbf{16.84}\cgreen{+24\%} & 15.07\cgreen{+11\%} \\
\addlinespace[3pt]

DROP
& 82.74\cred{-0.12} & \textbf{82.78}\cred{-0.08} & 82.70\cred{-0.16} \\
& 2.93 & \textbf{3.79}\cgreen{+29\%} & 3.15\cgreen{+8\%} \\
\addlinespace[3pt]

KorBench
& 33.20\cgreen{+0.08} & \textbf{35.04}\cgreen{+1.92} & 33.92\cgreen{+0.80} \\
& 3.72 & \textbf{5.03}\cgreen{+35\%} & 4.43\cgreen{+19\%} \\
\addlinespace[3pt]

HumanEval
& 34.15\cred{-0.61} & 36.59\cgreen{+1.83} & \textbf{37.80}\cgreen{+3.04} \\
& 3.82 & \textbf{4.69}\cgreen{+23\%} & 4.18\cgreen{+9\%} \\
\addlinespace[3pt]

LCB
& 8.15\cgreen{+0.00} & 7.54\cred{-0.61} & 7.71\cred{-0.44} \\
& 1.93 & \textbf{2.17}\cgreen{+12\%} & 2.00\cgreen{+4\%} \\
\addlinespace[3pt]

GSM8K
& \textbf{78.47}\cgreen{+0.53} & 77.18\cred{-0.76} & 77.48\cred{-0.46} \\
& 3.22 & \textbf{3.87}\cgreen{+20\%} & 3.39\cgreen{+5\%} \\
\addlinespace[3pt]

MATH
& 37.04\cred{-0.26} & \textbf{37.24}\cred{-0.06} & 37.18\cred{-0.12} \\
& 2.42 & \textbf{2.84}\cgreen{+17\%} & 2.55\cgreen{+5\%} \\
\midrule

\textbf{Average}
& 53.45\cred{-0.05} & 53.98\cgreen{+0.48} & \textbf{53.99}\cgreen{+0.49} \\
& 4.31 & \textbf{5.48}\cgreen{+27\%} & 4.77\cgreen{+11\%} \\
\bottomrule
\end{tabular}
}
\end{table}

\begin{table}[t]
\centering
\caption{
Main benchmark results \textbf{w/ Early Stop} across eight datasets on \textbf{LLaDA-MoE-Instruct}.
(Gen Length=256, Block Size=64). Cells show \textbf{Score} (top, relative to LLaDA-MoE) and \textbf{TPF} (bottom, improvement over Fast-dLLM).
}
\label{tab:Main-Results-MoE}
\renewcommand{\arraystretch}{0.95}
\setlength{\tabcolsep}{2.5pt}

\resizebox{0.95\columnwidth}{!}{
\begin{tabular}{lccc}
\toprule
\textbf{Benchmark} ${}^{^{\textbf{Score}}}_{_{\textbf{TPF}}}$ & \textbf{Fast-dLLM} & \textbf{CD} & \textbf{CD}$^{\dagger}$ \\
\midrule
MMLU
& 64.08\cgreen{+0.00} & \textbf{64.21}\cgreen{+0.13} & 63.94\cred{-0.14} \\
& 2.16 & \textbf{2.46}\cgreen{+14\%} & 2.33\cgreen{+8\%} \\
\addlinespace[3pt]

SQuAD2.0
& 86.88\cgreen{+0.00} & 87.27\cgreen{+0.39} & \textbf{87.35}\cgreen{+0.47} \\
& 7.09 & \textbf{9.64}\cgreen{+36\%} & 8.41\cgreen{+19\%} \\
\addlinespace[3pt]

DROP
& \textbf{80.16}\cgreen{+0.00} & 79.72\cred{-0.44} & 79.87\cred{-0.29} \\
& 2.73 & \textbf{3.28}\cgreen{+20\%} & 2.92\cgreen{+7\%} \\
\addlinespace[3pt]

KorBench
& \textbf{36.88}\cgreen{+0.16} & 36.48\cred{-0.24} & 36.64\cred{-0.08} \\
& 2.36 & \textbf{3.28}\cgreen{+38\%} & 2.73\cgreen{+16\%} \\
\addlinespace[3pt]

HumanEval
& 51.22\cgreen{+0.00} & 51.22\cgreen{+0.00} & \textbf{53.05}\cgreen{+1.83} \\
& 4.97 & \textbf{6.00}\cgreen{+21\%} & 5.45\cgreen{+10\%} \\
\addlinespace[3pt]

LCB
& 14.04\cgreen{+0.16} & 14.37\cgreen{+0.49} & \textbf{14.65}\cgreen{+0.77} \\
& 2.43 & \textbf{2.81}\cgreen{+16\%} & 2.55\cgreen{+5\%} \\
\addlinespace[3pt]

GSM8K
& 74.45\cgreen{+0.08} & \textbf{74.98}\cgreen{+0.61} & 74.37\cgreen{+0.00} \\
& 2.28 & \textbf{2.68}\cgreen{+18\%} & 2.42\cgreen{+6\%} \\
\addlinespace[3pt]

MATH
& 35.84\cred{-0.18} & \textbf{36.28}\cgreen{+0.26} & 36.26\cgreen{+0.24} \\
& 2.35 & \textbf{2.71}\cgreen{+15\%} & 2.48\cgreen{+6\%} \\
\midrule

\textbf{Average}
& 55.44\cgreen{+0.02} & 55.57\cgreen{+0.15} & \textbf{55.77}\cgreen{+0.35} \\
& 3.30 & \textbf{4.11}\cgreen{+25\%} & 3.66\cgreen{+11\%} \\
\bottomrule
\end{tabular}
}
\end{table}
In Section~\ref{sec:5.6}, we demonstrate the orthogonality and compatibility of CreditDecoding through experiments combining it with several acceleration techniques. Results show that CreditDecoding consistently improves both speed and performance across all tested methods.

In this section, we provide brief introductions to the acceleration methods discussed in Section~\ref{sec:5.6} and include additional TPS results to better illustrate CreditDecoding’s effectiveness, particularly on system-level accelerations that mainly improve TPS. We also extend our orthogonality analysis to LLaDA-MoE, with detailed results presented below.

We evaluate its orthogonality on four representative acceleration techniques, as illustrated in Figure~\ref{fig:Orthogonality}.

\textbf{Early Stop}: Early Stop terminates decoding when the current token is <EOS> and all previous tokens are finalized, effectively reducing redundant generation and improving decoding efficiency.

\textbf{Fast-dLLM}~\citep{wu2025fastdllmtrainingfreeaccelerationdiffusion}: A state-of-the-art acceleration method consisting of threshold-based parallel decoding and KV cache. Since the KV cache significantly increases TPS at the cost of performance, we mainly compare with Fast-dLLM (w/o KV).

\textbf{PyTorch Compiler}~\citep{pytorch20}: PyTorch Compiler leverages graph-level optimizations for runtime acceleration without altering decoding behavior. For MoE architectures, the compiler can fuse MoE kernels, which significantly accelerates the inference process.

\textbf{FP8 Quantization}~\citep{kwon2023vllm}: FP8 quantization is a technique that reduces the precision of floating-point numbers to 8-bit, aiming to accelerate deep learning models by lowering storage and computation costs while maintaining sufficient accuracy.

Table~\ref{tab:Orthogonality_Combined} presents detailed results corresponding to Figure~\ref{fig:Orthogonality}, including TPS comparisons and reports results under the same settings on LLaDA-MoE, further including FP8 quantization.

\subsection{Full-Distribution Trace Credit}
\label{Appendix:fullcredit}

In Sec.~\ref{sec:3.2}, for each position, trace credit is accumulated only on the top-$1$ candidate token at each step.
This focused strategy concentrates reinforcement and yields stronger acceleration, but it may ignore useful probability mass on other plausible candidates.

To examine the generality of credit accumulation beyond the top-$1$ hypothesis, we also consider a full-distribution variant that updates all tokens:
\begin{equation}
C_t^{i,v} = \beta C_{t+1}^{i,v} + (p_t^{i,v})^{\gamma}, \qquad \forall v \in \mathcal{V}.
\end{equation}
By aggregating historical evidence over the entire predictive distribution, this variant becomes less sensitive to transient noise and confidence oscillation. 
Intuitively, it can achieve slightly higher accuracy but weaker acceleration compared to the standard version, forming a more conservative trade-off option.

We evaluate this full-distribution variant (denoted as \textbf{CD}$^{\dagger}$) alongside the standard CreditDecoding (\textbf{CD}).
Tables~\ref{tab:Main-Results-8B} and~\ref{tab:Main-Results-MoE} report results on LLaDA-8B-Instruct and LLaDA-MoE-Instruct across eight benchmarks.

Overall, \textbf{CD}$^{\dagger}$ delivers slightly larger accuracy improvements than \textbf{CD}
(+$0.49$ vs. +$0.48$ on LLaDA-8B and +$0.35$ vs. +$0.15$ on LLaDA-MoE),
while incurring around $10\%$ lower speedup due to its weaker reinforcement strength.
Nevertheless, \textbf{CD}$^{\dagger}$ still achieves substantial acceleration,
showing that aggregating credit over the full predictive distribution remains effective relative to the baseline.

These results provide two important insights.
First, the comparable (and sometimes higher) accuracy of \textbf{CD}$^{\dagger}$ indicates that credit accumulation is fundamentally general and does not rely on reinforcing only the token that will be decoded.
Second, \textbf{CD} achieves stronger acceleration, indicating that assigning credit only to the top-1 token is sufficient. 
Since the target tokens typically maintains the highest confidence for several consecutive steps, this \textit{focused reinforcement} is both stable and computationally efficient.

Unlike \textbf{CD}, the full-distribution variant requires maintaining credits for every token in the vocabulary and updating them at each denoising step, which introduces non-trivial memory and computational overhead.
Therefore, \textbf{CD} provides a better efficiency–accuracy trade-off and remains the recommended default in practice.

\subsection{Robustness of the Focused Enhancement}
\label{Appendix:focus_robustness}

\begin{table}[t]
    \centering
    \caption{Performance and TPF across different credit enhancement strategies on top-$K$ candidate tokens and CD$^{\dagger}$, where top-$1$ denotes the default focus-enhancement mechanism applied in CD.}
    \begin{tabular}{lcc}
    \toprule
    \textbf{Enhancement Strategy} & \textbf{Score} & \textbf{TPF} \\
    \midrule
    Top-$1$ (CD) & 53.98 & \textbf{5.47} \\
    Top-$2$& 53.57 & 5.18 \\
    Top-$3$& 53.96 & 5.06 \\
    Top-$4$& 53.71 & 5.00 \\
    Top-$5$& 53.63 & 4.95 \\
    Top-$all$ (CD$^{\dagger}$) & \textbf{53.99} & 4.77 \\
    \bottomrule
    \end{tabular}
    \label{tab:topk_robustness}
\end{table}

For each denoising step, the focus enhancement mechanism strictly selects only the top-$1$ candidate token ranked by confidence to update the trace credit. To validate the robustness of this design choice, 
we evaluate a variant that updates trace credit for the top-$K$ candidate tokens ranked by confidence at each position, rather than only the top-$1$ token.

First, we investigate incorporating broader information into the credit matrix. As detailed in Appendix~\ref{Appendix:fullcredit}, we test a "Full-Distribution Trace Credit" approach, which records probabilities across all token positions regardless of their confidence levels. Furthermore, we conducted experiments using a \textit{top-$K$ enhancement strategy}, varying $K$ from 1 to 5. The results, along with the performance of our extended variant CD$^{\dagger}$, are summarized in Table~\ref{tab:topk_robustness}. 
Based on these results, we outline the following key observations:
\begin{itemize}
    \item Incorporating more tokens into the trace credit consistently degrades TPF. This occurs because the accumulated credit distribution becomes less "peaky," resulting in a more conservative denoising process.
    \item The score does not exhibit a reliable upward trend as $K$ increases. We hypothesize that increasing K introduces noisy or detrimental tokens that interfere with the denoising trajectory.
    \item Interestingly, as shown in Appendix~\ref{Appendix:fullcredit}, the CD$^{\dagger}$ variant achieves a slight performance gain (+0.2) on the LLaDA-MoE model. This suggests that tokens beyond the top-$1$ can theoretically assist denoising, provided they are rigorously filtered. Considering this alongside our second observation, we conclude that selecting tokens based simply on confidence rank may not be an effective strategy. 
\end{itemize}

Overall, updating the trace credit exclusively with the top-$1$ token remains the optimal configuration to balance inference efficiency and accuracy. Incorporating additional tokens sacrifices speed without guaranteeing stable performance improvements.

\subsection{End-to-End Efficiency}
\label{Appendix:end_to_end_efficiency}

To supplement the theoretical TPF metrics presented in the main text, we provide the detailed end-to-end inference speed, measured in Tokens Per Second (TPS), for all eight benchmarks on LLaDA-8B-Instruct. The evaluation is conducted under the same settings as Table~\ref{tab:Main-Results}.

\begin{table}[h]
\centering
\caption{
TPS comparison on \textbf{LLaDA-8B-Instruct} (Gen Length=256, Block Size=64). 
For our method, the cells show the absolute \textbf{TPS}, along with the speedup multipliers over the Baseline (superscript) and Fast-dLLM (subscript).
}
\label{tab:tps_results}
\renewcommand{\arraystretch}{1.75}
\setlength{\tabcolsep}{4pt}

\resizebox{0.99\columnwidth}{!}{
\begin{tabular}{lccc}
\toprule
\textbf{Benchmark} & \textbf{Baseline} & \textbf{Fast-dLLM} & \textbf{CD} ${}^{\text{vs. Baseline}\vphantom{\big|}}_{\vphantom{\small|}\text{vs. Fast-dLLM}}$ \\
\midrule
MMLU
& 4.95 & 12.49 & \textbf{17.30}$^{\textcolor{gray}{+249\%}\vphantom{\small|}}_{\vphantom{\small|}\textcolor{green!50!black}{+38.4\%}}$ \\

SQuAD2.0
& 1.61 & 10.57 & \textbf{12.94}$^{\textcolor{gray}{+706\%}\vphantom{\small|}}_{\vphantom{\small|}\textcolor{green!50!black}{+22.4\%}}$ \\

DROP
& 5.61 & 15.02 & \textbf{18.95}$^{\textcolor{gray}{+238\%}\vphantom{\small|}}_{\vphantom{\small|}\textcolor{green!50!black}{+26.2\%}}$ \\

KorBench
& 6.33 & 12.40 & \textbf{14.56}$^{\textcolor{gray}{+130\%}\vphantom{\small|}}_{\vphantom{\small|}\textcolor{green!50!black}{+17.4\%}}$ \\

HumanEval
& 16.04 & 59.25 & \textbf{67.69}$^{\textcolor{gray}{+322\%}\vphantom{\small|}}_{\vphantom{\small|}\textcolor{green!50!black}{+14.2\%}}$ \\

LCB
& 4.55 & 8.58 & \textbf{9.54}$^{\textcolor{gray}{+110\%}\vphantom{\small|}}_{\vphantom{\small|}\textcolor{green!50!black}{+11.2\%}}$ \\

GSM8K
& 18.84 & 58.37 & \textbf{69.95}$^{\textcolor{gray}{+271\%}\vphantom{\small|}}_{\vphantom{\small|}\textcolor{green!50!black}{+19.8\%}}$ \\

MATH
& 18.20 & 43.35 & \textbf{50.81}$^{\textcolor{gray}{+179\%}\vphantom{\small|}}_{\vphantom{\small|}\textcolor{green!50!black}{+17.2\%}}$ \\
\midrule

\textbf{Average}
& 9.52 & 27.50 & \textbf{32.72}$^{\textcolor{gray}{+244\%}\vphantom{\small|}}_{\vphantom{\small|}\textcolor{green!50!black}{+19.0\%}}$ \\
\bottomrule
\end{tabular}
}
\end{table}

As shown in Table~\ref{tab:tps_results}, CreditDecoding consistently improves TPS across all benchmarks, achieving an average gain of $+19.0\%$ over Fast-dLLM and $+244\%$ over the baseline. These TPS results further demonstrate that the efficiency improvements indicated by TPF translate directly into significant end-to-end speedups, confirming the practical efficacy of CreditDecoding in real-world deployments.

\subsection{Failure Mode Analysis}
\label{Appendix:failure_mode}

While CreditDecoding maintains or improves performance on datasets with high parallel decoding potential, we observe slight performance drops on a few reasoning-intensive tasks, as shown in Table~\ref{tab:Main-Results}, such as mathematical reasoning and complex code generation. These datasets exhibit lower TPF and require longer denoising trajectories. In contrast, all datasets with $TPF > 4 $ maintain or improve performance. We attribute this to the following factors.

\textbf{Strong Causal Dependency.} As illustrated in Figure~\ref{fig:gsm_confdc}, the predictive distribution for later tokens in these tasks is highly dependent on preceding ones, resulting in high entropy. This forces the model to behave in a strictly autoregressive-like manner during denoising. Consequently, when the context is not yet fully formed—especially under aggressive parallel decoding—prediction errors in earlier steps are easily propagated and amplified for later tokens.
    
\textbf{Unstable Confidence Convergence.} As shown in Figure~\ref{fig:conf_convergence}, tokens requiring a larger number of total denoising steps (the yellow curve) exhibit significantly less stable confidence convergence compared to those with higher TPF (the blue curve). When the temporal consistency of confidence weakens, the credit accumulation process struggles to effectively distinguish true convergence signals from noise under fixed hyperparameter configurations.

Nevertheless, we find that this vulnerability is fundamentally tied to the intrinsic capability of the base model. As shown in Table~\ref{tab:Main-Results}, these performance drops are significantly mitigated on the stronger model \textit{LLaDA-MoE}. This suggests that more capable models exhibit stronger and more stable temporal consistency during the denoising trajectory, making the trace credit accumulation process inherently more robust to complex reasoning tasks.

Furthermore, the hyperparameters in Table~\ref{tab:Main-Results} were selected to maximize average score across all eight datasets. While this may lead to suboptimal performance on certain reasoning-heavy edge cases, it establishes a robust baseline for general dLLM acceleration. Future work could explore dynamic or task-adaptive hyperparameter scheduling to further bridge the gap in these specialized domains.
\end{document}